\documentclass[sigconf,nonacm]{acmart}
\graphicspath{{Figures/}}
\usepackage{booktabs}
\usepackage{enumitem}
\usepackage{graphicx}
\usepackage{amsmath}
\usepackage{colortbl}

\title{CastWorld: The World Behind the Numbers — Benchmarking Context-Aware Agentic Time Series Forecasting}

\author{Bokai Pan}
\affiliation{%
  \institution{State Key Laboratory of Cognitive Intelligence, University of Science and Technology of China}
  \city{Hefei, Anhui Province}
  \country{China}
}
\email{bkpan@mail.ustc.edu.cn}

\author{Ruijie Yang}
\affiliation{%
  \institution{State Key Laboratory of Cognitive Intelligence, University of Science and Technology of China}
  \city{Hefei, Anhui Province}
  \country{China}
}
\email{yangruijie@mail.ustc.edu.cn}

\author{Mingyue Cheng}
\affiliation{%
  \institution{State Key Laboratory of Cognitive Intelligence, University of Science and Technology of China}
  \city{Hefei, Anhui Province}
  \country{China}
}
\email{mycheng@ustc.edu.cn}

\author{Jiahao Wang}
\affiliation{%
  \institution{State Key Laboratory of Cognitive Intelligence, University of Science and Technology of China}
  \city{Hefei, Anhui Province}
  \country{China}
}
\email{jiahao.wang@mail.ustc.edu.cn}

\author{Xiaoyu Tao}
\affiliation{%
  \institution{State Key Laboratory of Cognitive Intelligence, University of Science and Technology of China}
  \city{Hefei, Anhui Province}
  \country{China}
}
\email{txytiny@mail.ustc.edu.cn}

\AtEndPreamble{%
  \hypersetup{hidelinks}%
}

\begin{document}
\setlength{\emergencystretch}{1.2em}

\begin{abstract}
Agentic time series forecasting extends conventional approaches beyond one-shot numerical prediction toward autonomous, context-aware interaction.
However, existing benchmarks offer limited integrated support for multidimensional real-world context modeling, a standardized forecasting harness, and systematic evaluation of agentic forecasting.
In this paper, we introduce CastWorld, an agentic benchmark for context-aware time series forecasting.
First, we enrich 25 real-world datasets across eight domains with exogenous variables, events, domain knowledge, and historical cases using a construction pipeline that jointly assesses scientific validity and predictive utility.
Second, a standardized forecasting harness integrates 49 core tools, domain-specific tools, and a state-adaptive loop. The loop tracks forecast versions, allowing intermediate forecasts to guide evidence gathering and new evidence to inform revision and final selection.
Third, model-level evaluation assesses forecast quality, process execution, behavioral trustworthiness, and resource efficiency through decision trajectories and forecast-version histories.
Also, system-level evaluation compares predictive accuracy across forecasting systems. Experiments reveal that most models exhibit limited capacity for continuous context acquisition during iterative forecasting. Deficiencies in forecasting tool selection and use constrain predictive gains. Forecast self-evaluation is also insufficient for consistently identifying and submitting the best generated forecast version. CastWorld provides a unified foundational platform for evaluating the entire agentic time series forecasting process in real-world environments
~\footnote{\url{https://github.com/CastWorld-Bench/CastWorld}}.
\end{abstract}

\vspace{-10pt}
\keywords{Agentic time series forecasting, Context-aware forecasting, Real-world context modeling, Interactive benchmarks}
\maketitle
\vspace{-10pt}
\section{Introduction}
\label{sec:introduction}

Time series forecasting often requires interpreting temporal patterns in light of external conditions, events, and domain knowledge~\cite{liu2024timemmd,williams2025cik}. Exogenous variables and events characterize \mbox{real-world} conditions, while domain knowledge and historical cases describe the dynamics underlying observed time series. Recent large language model (LLM)-driven agents combine contextual reasoning with numerical tools to construct and revise forecasts~\cite{tao2026castr1,tao2026castfsr}. This interactive process involves deciding what evidence to acquire, which tools to use, and how to revise predictions as evidence accumulates. Studying these capabilities calls for a shared forecasting environment that represents the relevant real-world context and supports autonomous investigation throughout forecast formation.

\begin{figure}[t]
  \centering
  \includegraphics[width=\columnwidth]{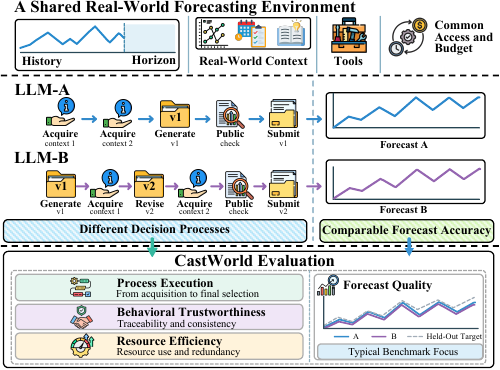}
  \vspace{-20pt}
  \caption{Motivation of CastWorld. Different decision processes can get comparable accuracy in shared real-world contexts, motivating joint evaluation of outcomes and strategies.}
  \Description{A shared forecasting task provides history, a horizon, contextual information and tools under common access and budget constraints. LLM-A acquires two context observations before generating, checking and submitting version v1. LLM-B generates v1, acquires context, revises to v2, acquires further context, checks and submits v2. These are illustrative decision traces under the same environment, not prescribed workflows. Their similar but distinct blue and violet forecasts are compared with a gray dashed held-out target offline. Different decision processes map to process execution, behavioral trustworthiness and resource efficiency, while comparable accuracy maps to forecast quality, marked Typical benchmark focus. All four dimensions belong to CastWorld evaluation. Public checks never use errors on held-out targets.}
  \label{fig:motivation}
  \vspace{-12pt}
\end{figure}

Existing context-aware forecasting benchmarks have advanced evaluation from context utilization to evidence discovery. Time-MMD aligns numerical series with domain-specific textual information~\cite{liu2024timemmd}. Context is Key (CiK) constructs tasks requiring supplied textual context~\cite{williams2025cik}, while Dr-CiK evaluates evidence discovery and synthesis from noisy corpora for downstream forecasting~\cite{tang2026drcik}. A further research question is how effectively models form and refine forecasts through continued investigation of the real-world context surrounding a target series. An intermediate prediction may reveal an information gap, prompting further evidence gathering, while newly acquired evidence may guide subsequent forecast revision and final selection. Figure~\ref{fig:motivation} illustrates a shared forecasting environment that combines historical observations, real-world context, and forecasting tools under the same access rules and budget. Models operating in this environment can achieve comparable forecast accuracy through different decision processes, motivating systematic evaluation of the full forecasting process.

Despite this progress, three limitations remain in existing benchmarking environments. First, existing resources provide limited support for multidimensional real-world context modeling that captures both environmental conditions and system dynamics across forecasting tasks and domains. Second, existing benchmarks offer limited support for a standardized forecasting harness that makes contextual resources accessible through modular tools and supports state-adaptive interactions between evidence acquisition and forecast refinement. Third, systematic evaluation of the complete forecasting process remains limited, particularly in linking evidence acquisition and tool use to successive forecast versions, final selection, and resource costs.

\begin{figure*}[t]
  \centering
  \includegraphics[width=\textwidth]{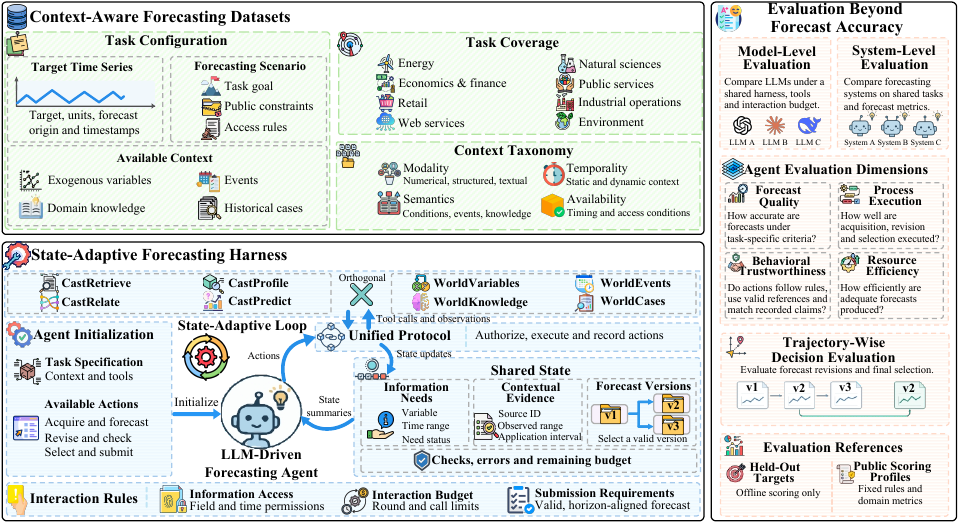}
  \vspace{-15pt}
  \caption{Overview of CastWorld. Context-aware datasets model real-world forecasting environments. A standardized harness organizes tools by context and capability and supports state-adaptive interaction. Decision trajectories and forecast-version histories support four-dimensional model evaluation, complemented by system-level accuracy comparisons.}
  \Description{The figure has three regions. The upper-left region organizes forecasting tasks, domain coverage, contextual resources and the four context classification axes. The lower-left region shows the forecasting harness, including agent initialization, four tool families, interaction rules and a state-adaptive loop. Agent actions pass through a unified protocol, which authorizes, executes and records them. Shared state contains information needs, contextual evidence, forecast versions, checks, errors and remaining budget. State summaries inform subsequent decisions. The right region distinguishes model-level evaluation under a shared harness from system-level forecast comparisons. It presents four agent evaluation dimensions, trajectory-wise evaluation of revisions and final selection, and evaluation references comprising held-out targets and public scoring profiles. Arrows illustrate interactions within the harness. Forecast curves and version identifiers are schematic.}
  \label{fig:overview}
  \vspace{-8pt}
\end{figure*}

To address these limitations, we introduce CastWorld, an agentic benchmark for context-aware time series forecasting that also supports Temporal World Model research. Figure~\ref{fig:overview} summarizes its design. CastWorld enriches 25 real-world datasets across eight domains with exogenous variables, events, domain knowledge, and historical cases through a construction pipeline that jointly assesses scientific validity and predictive utility. Its lightweight, modular forecasting harness provides standardized interfaces to 49 core tools and domain-specific extensions, with a state-adaptive loop supporting evidence acquisition, forecast construction, checking, revision, and final selection. Model-level evaluation connects execution trajectories and forecast-version histories with subsequently observed outcomes, assessing forecast quality, process execution, behavioral trustworthiness, and resource efficiency. System-level evaluation compares predictive accuracy across forecasting systems. Experiments reveal that most models exhibit limited capacity for continuous context acquisition during iterative forecasting. Deficiencies in forecasting tool selection and use constrain predictive gains. Forecast self-evaluation is also insufficient for consistently identifying and submitting the best forecast version. These findings identify key capability gaps in translating real-world context and forecasting tools into reliable predictions.

Our main contributions are summarized as follows:
\begin{itemize}
    \item Datasets with multidimensional real-world context.
    CastWorld enriches 25 datasets across eight domains through a construction pipeline that integrates four complementary types of context and jointly assesses their scientific validity and predictive utility.

    \item A standardized and extensible forecasting harness.
    We develop a modular harness integrating 49 core tools and domain-specific extensions through standardized interfaces, with a state-adaptive loop for evidence acquisition, forecast revision, and final selection.

    \item Decision-centric evaluation and empirical findings.
    We establish model-level evaluation across four dimensions and system-level comparisons of predictive accuracy. Experiments reveal limitations in most models' continuous context acquisition, effective forecasting tool use, and forecast self-evaluation and version selection.
\end{itemize}

\section{Related Work}
\label{sec:related-work}

\subsection{Time Series Forecasting Methods}

Statistical forecasting models explicitly characterize temporal dependencies and exogenous effects~\cite{pankratz1991dynamic}, while deep learning methods learn forecasting functions using architectures ranging from linear models to Transformers~\cite{cheng2026tsfsurvey,zeng2023dlinear,nie2023patchtst}.
Time series foundation models leverage large-scale pretraining across diverse datasets to support generalization across forecasting tasks~\cite{das2024timesfm,ansari2024chronos}.
Beyond numerical modeling, LLM-based methods incorporate language understanding and contextual reasoning into forecasting~\cite{gruver2023llmtime,requeima2024llmprocesses,cheng2026timereasoner}, leveraging exogenous signals and historical analogues to provide additional evidence for predictions~\cite{huang2025exollm,ma2026quizsf,jalori2025flairrts}.
Building on these capabilities, agentic frameworks coordinate evidence acquisition, tool use, and iterative forecast refinement during inference~\cite{zhang2025alphacast,ye2026tsreasoner,tao2026castr1,tao2026castfsr}.
Rather than generating predictions in a single step, these frameworks allow agents to gather relevant information, invoke forecasting tools, and revise intermediate results as new evidence becomes available~\cite{cheng2026atsf}.
Consequently, forecasting in agentic settings involves not only predictive accuracy but also decisions about information acquisition, forecast revision, and final prediction selection.

\subsection{Context-Aware Forecasting Benchmarks}

Existing forecasting benchmarks address different aspects of predictive evaluation. Conventional broad-coverage suites provide comparisons across datasets, forecast horizons, and model adaptation settings~\cite{godahewa2021monash,qiu2024tfb,zhang2024probts,li2025tsfmbench}. Contextual and reasoning benchmarks assess how external text and events support forecasting and evaluate temporal reasoning quality~\cite{liu2024timemmd,williams2025cik,liu2026world2ts,weng2026temporalbench,ahamed2026tfrbench}. GIFT-Eval-CTX studies how verified future effects inform numerical forecasts, including predictions generated by tool-using agents~\cite{liu2026giftctx}. Interactive benchmarks investigate how information acquisition and tool use support forecasting decisions. Dr-CiK focuses on iterative evidence discovery from noisy corpora~\cite{tang2026drcik}. FW-Bench examines how budget-constrained choices of forecasting models, historical windows, and forecast horizons affect downstream capacity planning and costs~\cite{nagashima2026fwbench}. These benchmarks cover complementary aspects of forecasting from predictive performance and contextual reasoning to interactive evidence acquisition and downstream operational decisions. Their task formulations and evaluation objectives reflect the capabilities and application requirements under investigation.

\subsection{Interactive Agent Evaluation}

Interactive agent benchmarks evaluate task completion through actions and environmental feedback across diverse task settings. General-purpose benchmarks cover tool use and Web navigation~\cite{liu2024agentbench,zhou2024webarena}, while stateful tool-use benchmarks~\cite{lu2025toolsandbox}
emphasize dependencies across actions and compliance with environment-specific constraints~\cite{yao2025taubench}.
In time-series analysis, specialized benchmarks address engineering tasks and constrained workflows~\cite{cai2025timeseriesgym,ye2025tsaia}, as well as multi-turn analytical tasks with evolving states~\cite{kong2026timesagemt,yao2026timesageev}. Other evaluation frameworks examine intermediate progress~\cite{ma2024agentboard}, evidence acquisition and research outcomes~\cite{chen2026trace,wang2025paperarena}, and reproducibility and accuracy-cost trade-offs~\cite{kapoor2026hal}. These studies cover complementary aspects of agent performance, but vary in their evaluation of intermediate decisions and final outcomes. Task and outcome validity remain essential for ensuring that benchmark results reliably reflect the capabilities under study~\cite{zhu2025abc}.

\section{Benchmark Datasets}
\label{sec:evaluation-datasets}
CastWorld constructs a multidimensional representation of the real-world environment surrounding each target series. Exogenous variables and events characterize environmental conditions and changes, while domain knowledge and historical cases describe operating mechanisms and past responses. Together, these resources form a dataset-level information space in which agents can investigate the conditions and dynamics relevant to their forecasts.

\begin{figure*}[t]
  \centering
  \includegraphics[width=\textwidth]{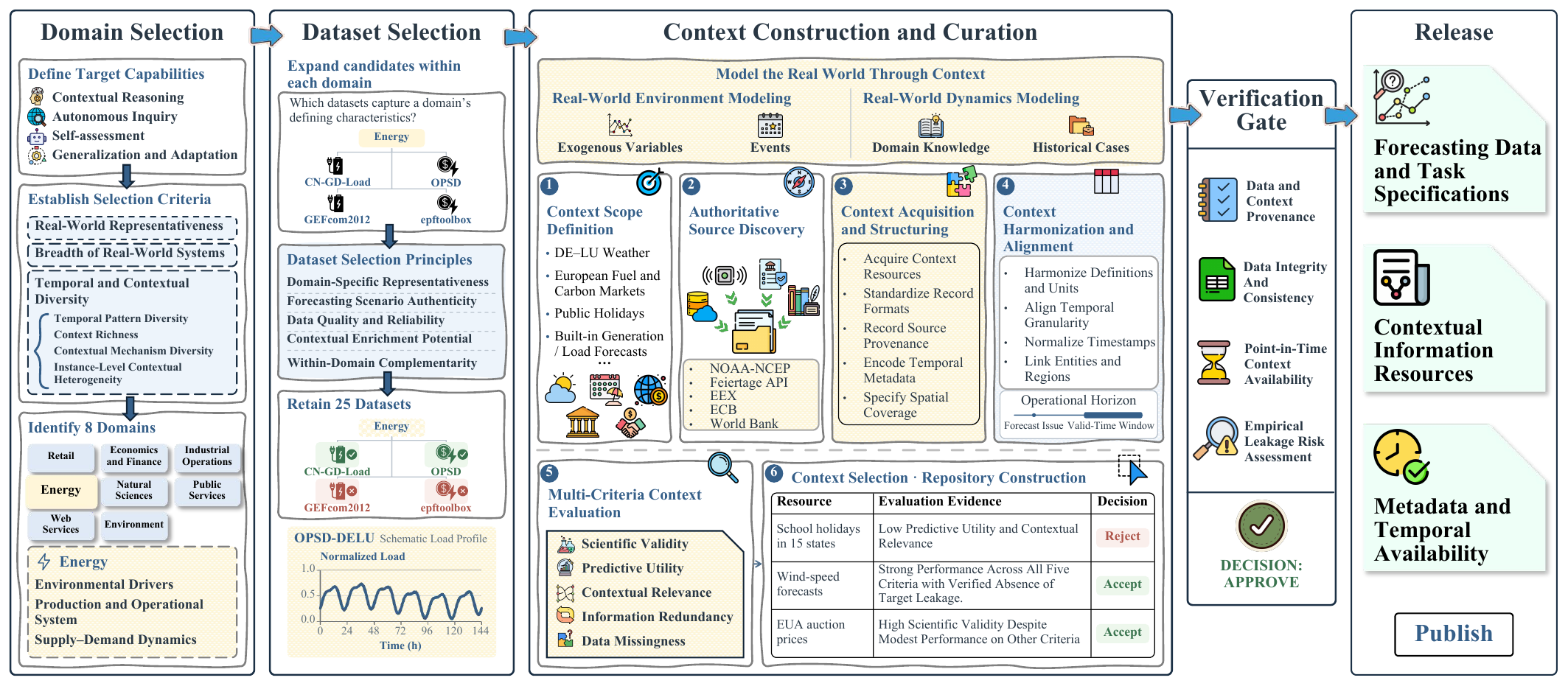}
  \caption{CastWorld's dataset construction pipeline. Domain and dataset selection establish representative forecasting scenarios. Four types of context capture environmental conditions and system dynamics, with scientific validity and predictive utility jointly guiding curation before verification and release.}
  \Description{A left-to-right pipeline with five stages: domain selection, dataset selection, context construction and curation, verification, and release. Context comprises exogenous variables, events, domain knowledge, and historical cases. Six curation steps cover scope definition, source discovery, acquisition, alignment, evaluation, and repository construction.}
  \label{fig:data-construction}
\end{figure*}

\subsection{Dataset Design Principles}
Domain selection emphasizes real-world representativeness, breadth across natural and sociotechnical systems, and temporal and contextual diversity. We seek complementary temporal patterns, heterogeneous contextual resources, different mechanisms linking context to the target, and variation in the information required by individual forecasting instances. This coverage supports the evaluation of contextual reasoning, autonomous inquiry, self-assessment, and generalization and adaptation.

Within each domain, datasets are selected for representative real-world forecasting scenarios, verifiable data quality, traceable sources and processing, and the potential to integrate external context. We also consider how datasets complement one another in their targets, temporal patterns, and driving mechanisms, so that the collection captures meaningful variation within each domain.

At the context level, selection jointly considers scientific validity and predictive utility, alongside contextual relevance, information redundancy, and data missingness. Scientific validity concerns the evidence linking a resource to the target and the conditions under which that relationship applies; predictive utility concerns its contribution under the task's information conditions, including information availability at the forecast origin. Together, these criteria guide the assembly of scientifically grounded and complementary contextual resources.

\begin{figure}[t]
  \centering
  \includegraphics[width=\columnwidth]{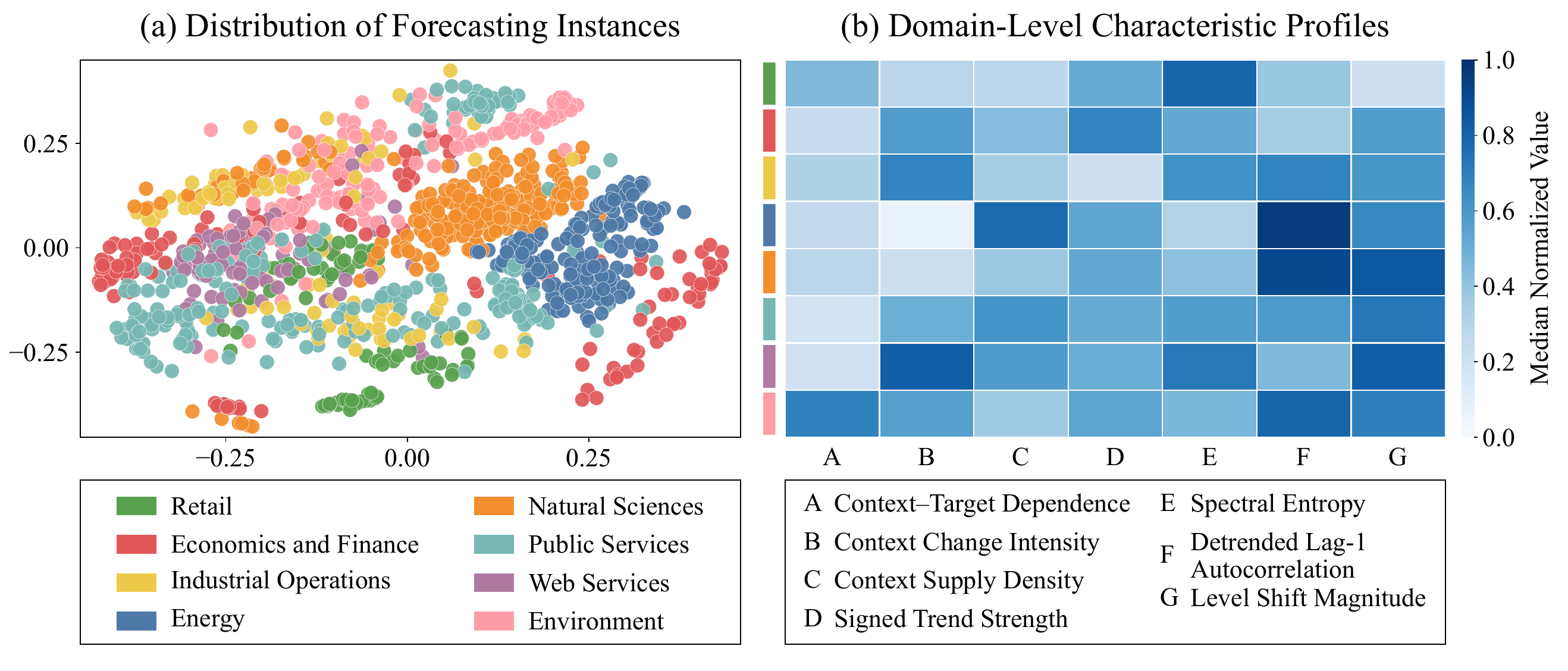}
  \vspace{-20pt}
  \caption{Task coverage across eight application domains. Domain colors link instances to profiles. A--G denote normalized contextual and temporal features.}
  \Description{Forecasting instances are colored by application domain. Matching row colors link the domains to a heatmap of seven normalized contextual and temporal characteristics, labeled A through G with full names below.}
  \label{fig:data-profiles}
  \vspace{-10pt}
\end{figure}

\subsection{Dataset Construction Pipeline}
Figure~\ref{fig:data-construction} summarizes the pipeline from domain and dataset selection to context curation, verification, and release. For each selected dataset, the target definition, spatial and temporal coverage, and relevant mechanisms guide the search for context in authoritative databases, public agencies, and domain literature. Exogenous variables are organized as numerical or categorical fields, and events as dated records. Literature-based knowledge curation extracts supported findings, checks their applicability, consolidates overlapping entries, and reviews conflicting claims, producing self-contained entries with source references and conditions of use. Historical cases pair earlier trajectories with their observed continuations, with each complete case ending before the training cutoff.

A common specification describes each resource's modality, temporal behavior, forecasting role, and availability. We reconcile definitions, units, normalization, aggregation, time zones, and temporal granularity, and link records to the relevant entities and regions. Temporal alignment distinguishes event occurrence from publication, and forecast issue times from valid-time windows. At the field level, statistical, deep learning, and time series foundation models compare history-only forecasts with forecasts using individual contextual fields. Random masking and reconstruction inform missing-value treatment, reconstruction from other fields diagnoses redundancy, and association analysis considers lags and trends. Manual review combines these empirical findings with source evidence and domain knowledge to determine resource inclusion and permitted access horizons, forming a dataset-level context repository.

Before release, verification checks data and context provenance, integrity and consistency, temporal availability, and leakage risks. Checks cover completeness, time ordering, and alignment of public and private records; unusually large predictive gains prompt inspection for target overlap or retrospective information. Resource access follows temporal metadata and declared task assumptions, including perfect-forecast settings where specified (Appendix~\ref{sec:task-profiles}). The release brings together forecasting data and task specifications, contextual resources, and metadata with temporal access conditions. Each episode specifies target history, forecast timestamps and horizon, training cutoff, resource catalogs, and access and submission constraints. Task manifests retain episode identifiers and source and configuration hashes.

\subsection{Dataset Statistics}
The suite contains 1,310 forecasting episodes from 25 datasets across eight domains, with sampling frequencies ranging from five minutes to monthly. Its dataset-level repositories contain 1,220 contextual fields, 1,520 knowledge entries, and 2,308 historical cases, averaging 48.8 fields and 60.8 knowledge entries per dataset. The knowledge libraries cover variable definitions, domain relationships, and conditions of use. Appendix~\ref{sec:dataset-resources} provides the complete inventory of task specifications, contextual resources, and access conditions.

Contextual complexity varies with information sources, spatial and temporal relationships, and resource availability. For example, iFLYTEK-Wind combines weather products across measurement heights, NYC-311 aligns weather with scheduled activities across city boroughs, and CAMELS-SPAT provides upstream and downstream observations for examining lagged hydrological responses. Domain knowledge describes relevant relationships and their conditions of use, while historical cases supply earlier trajectories and observed continuations. These resources represent different real-world forecasting environments. Figure~\ref{fig:data-profiles} complements this characterization with instance-level variation and domain-level profiles of context--target dependence, context change intensity, context supply density, and temporal dynamics.

For model-level evaluation, we select a fixed set of ten episodes per dataset to balance dataset coverage with inference costs and evaluation time. Selection uses offline target-curve descriptors, contextual and temporal coverage, and primary-context completeness. All 15 LLMs are evaluated on the same 250 episodes under identical task specifications and scoring profiles. Selection is independent of model results, and the descriptors remain unavailable to agents.

\section{Agentic Forecasting Harness}
\label{sec:harness}
CastWorld provides a standardized and extensible harness for autonomous, context-aware forecasting. Its tools, state-adaptive loop, and state management connect investigation of real-world context with forecast construction, checking, revision, and final selection.

\subsection{Forecasting Tools and Actions}
The harness provides 49 core tools and domain-specific extensions. We organize context-aware tool use along two orthogonal axes. The capability axis comprises CastRetrieve for evidence acquisition, CastProfile for descriptive analysis, CastRelate for relating context to target behavior, and CastPredict for forecast construction and revision. The context axis comprises WorldVariables, WorldEvents, WorldKnowledge, and WorldCases, corresponding to exogenous variables, events, domain knowledge, and historical cases. This context--capability matrix connects the datasets' contextual resources with the operations through which agents investigate the forecasting environment. General-purpose tools support analysis across domains, while domain-specific extensions provide specialized diagnostics.

\begin{figure}[t]
    \centering
    \includegraphics[width=\linewidth]{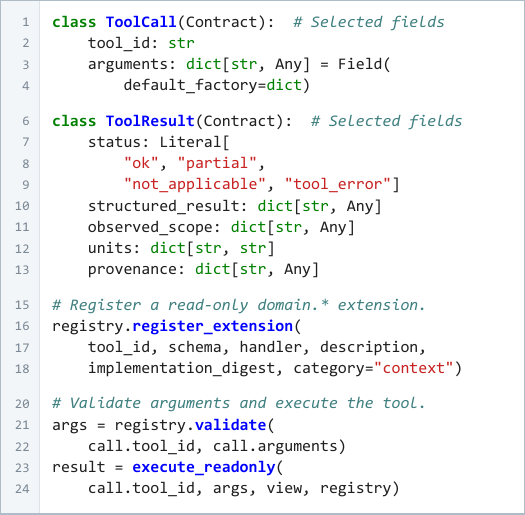}
    \caption{Standardized tool interface. Selected contract fields
    and interface calls illustrate validation, structured
    observations, and registration of read-only domain extensions.}
    \label{fig:tool_interface}
\end{figure}

Figure~\ref{fig:tool_interface} illustrates the standardized tool interface. A shared registry exposes tool descriptions and typed argument schemas. The runtime validates arguments before execution and returns read-only results through a common observation contract containing structured outputs, observed scopes, units, and provenance. Domain extensions bind unique identifiers to argument schemas and execution handlers, with implementation digests recorded in the registry. This design allows domain-specific analysis to extend the forecasting environment while reusing the same invocation, observation, and state-management mechanisms.

Through the shared action interface, agents declare information needs, invoke tools, register evidence, construct or revise forecasts, request public checks, and select a version for submission. Native candidates become agent-created forecast versions before submission, allowing tool-generated predictions to be traced through subsequent revision and final selection.

\subsection{State-Adaptive Forecasting Loop}
At each round $r$, the agent selects an action based on its authorized view of the current state. The harness validates and executes permitted operations, updates the state, and returns observations and feedback. Intermediate forecasts can reveal information gaps and guide further evidence acquisition, while new evidence can guide forecast revision and final selection. Public-check results and execution errors can also inform subsequent actions. Each ordinary interaction round consumes one budget unit, $b_{r+1}=b_r-1$, and its read-only calls $\mathcal{A}^{\mathrm{read}}_r$ satisfy $|\mathcal{A}^{\mathrm{read}}_r|\leq C_{\max}$, where $C_{\max}$ is the per-round read-only call limit. Within this budget, agents can adapt the order of actions, acquire further evidence after forecasting, revise or revisit earlier versions, and submit when ready. An episode ends with a valid submitted version or a recorded failure.

\subsection{State Management Mechanism}
The harness maintains the structured interaction state:
\begin{equation}
S_r=(N_r,O_r,E_r,V_r,C_r,b_r),
\label{eq:castworld_state}
\end{equation}
where $N_r$, $O_r$, $E_r$, $V_r$, and $C_r$ represent information needs, acquired observations, registered evidence, forecast versions, and execution feedback, with $b_r$ denoting the remaining budget. Together, these records track evidence acquisition, forecast versions and public checks, and action execution with errors and budget use. Agents receive bounded state summaries and can retrieve complete records by identifier. The observation history is retained when the active forecast changes, allowing agents to revisit earlier versions while retaining access to previously acquired information. Schema, state-version, and permission checks govern execution.

Evidence records link acquired observations and their source scopes to the forecast intervals where they are applied, while agent-generated rationales are recorded separately. Each accepted forecast construction or revision creates an immutable version:
\begin{equation}
v_k=(\hat{\mathbf{y}}^{(k)},p_k,o_k,E_k,r_k),
\label{eq:forecast_version}
\end{equation}
where $\hat{\mathbf{y}}^{(k)}$ gives forecast values, $p_k$ source-version references, $o_k$ the construction or revision operation, $E_k$ evidence references, and $r_k$ the creation round. Source-version references identify the forecasts used by an operation, while the preceding active version provides the baseline for checking declared numerical changes. For the first agent-created version, this comparison uses an operation-defined baseline when available. Decision trajectories and forecast-version histories thus link acquired evidence, numerical revisions, and final selection for offline evaluation. Held-out targets and offline version scores remain unavailable to the agent.

\section{Experimental Settings}
\label{sec:experimental-setup}
Our primary model-level evaluation compares LLM-driven agents under a shared forecasting harness with the same tasks, tools, access permissions, interaction budget, and scoring protocol. System-level evaluation compares predictive accuracy across forecasting systems on shared tasks.

\subsection{Evaluation Protocol and Metrics}
\label{sec:evaluation}
Controlled experiments vary contextual resources, tools, state summaries, interaction budgets, and access to new evidence during refinement. Comparisons matched on MASE and RMSSE examine decision behavior at comparable accuracy, while forecast-version histories trace numerical revisions and final selection. Model-level evaluation assesses forecast quality ($F$), process execution ($P$), behavioral trustworthiness ($T$), and resource efficiency ($E$) using numerical metrics and rule-based checks on forecasts, observations, and decision trajectories, without an LLM judge. We report all four dimensions alongside their weighted total to characterize predictive performance, forecasting behavior, and resource use.

\subsubsection{Forecast quality.}
For horizon $H$, let $y_h^{+}$ and $\hat{y}_h$ denote the held-out target and submitted prediction at step $h$. Forecast accuracy is measured by MASE and RMSSE, normalized using scales derived from training observations:
\begin{equation}
\mathrm{MASE}=
\frac{\sum_{h=1}^{H}|y_h^{+}-\hat{y}_h|}
{H s_{\mathrm{abs}}},
\quad
\mathrm{RMSSE}=
\sqrt{\frac{\sum_{h=1}^{H}(y_h^{+}-\hat{y}_h)^2}
{H s_{\mathrm{sq}}}},
\label{eq:forecast_errors}
\end{equation}
where $s_{\mathrm{abs}}$ and $s_{\mathrm{sq}}$ are the mean absolute and mean squared training differences at the first valid lag, testing the seasonal lag before lag one. If neither lag yields valid scales from sufficient finite pairs, the fallback scale is the maximum of the training interquartile range, absolute median, and a unit floor; its square is used for $s_{\mathrm{sq}}$. The base score is
$F_{\mathrm{base}}=100/[1+(0.45\,\mathrm{MASE}
+0.35\,\mathrm{RMSSE})/0.80]$.
For tasks with additional requirements, such as peak amplitude, ramp behavior, or window totals, the evaluation incorporates the corresponding normalized losses $\tilde{\ell}_j$:
\begin{equation}
F=\left(1-\sum_j w_j\right)F_{\mathrm{base}}
+\sum_j w_j\frac{100}{1+\tilde{\ell}_j},
\label{eq:forecast_score}
\end{equation}
where $w_j$ denotes the weight of each additional component. The normalization procedure and scoring profile are fixed for each task before evaluation.

\subsubsection{Process execution.}
The process score $P$ evaluates valid actions, alignment between information requests and declared needs, evidence support for numerical revisions, revision control, active checking, and final version selection. Their respective weights are 20\% each for valid actions, need--query alignment, and revision support; 15\% each for revision control and final selection; and 10\% for checking. Revision support checks whether referenced evidence or scoped rationales cover the changed forecast intervals. Active checking evaluates whether agents use public checks to inspect their forecasts before submission. Final-selection quality compares the submitted forecast with the best agent-created version according to the offline forecast-quality score $F$. This check applies when at least two agent-created versions exist; their offline scores and rankings remain hidden during interaction.

\subsubsection{Behavioral trustworthiness.}
The behavioral trustworthiness score $T$ measures output validity, permission compliance, reference integrity, satisfaction of public constraints, consistency between declared and executed operations, and error recovery. The first three criteria each receive 20\% weight, public constraints and declared--executed consistency receive 15\% each, and error recovery receives 10\%. For $P$ and $T$, applicable checks are aggregated with their weights renormalized. Applicability follows each check's requirements; for example, revision-specific checks apply to trajectories containing numerical revisions.

\subsubsection{Resource efficiency.}
Resource efficiency accounts for execution time, token consumption, and redundant operations. Time and token scores are normalized against fixed validation anchors:
\begin{equation}
R_z=\frac{100}{1+\bar{z}/z_0},
\quad z\in\{t,k\},
\end{equation}
where $\bar{z}$ is the mean execution time or token consumption across the planned episodes for a given model, condition, and dataset, and $z_0$ is the validation anchor. Redundancy is scored as follows:
\begin{equation}
R_b=100\left[1-
\frac{r_{\mathrm{redundant}}+r_{\mathrm{repeat}}}{2}\right],
\quad
C=0.4R_t+0.4R_k+0.2R_b,
\end{equation}
where $r_{\mathrm{redundant}}$ and $r_{\mathrm{repeat}}$ measure the rates of redundant operations and repeated errors, respectively. To account for forecast adequacy, let $F^0$ denote the score of a fixed history-only reference forecast under the same task scoring profile. We define:
\begin{equation}
a=\operatorname{clip}(2-F^0/F,0,1),
\qquad
E=70a+(0.6-0.3a)C,
\label{eq:efficiency_score}
\end{equation}
with $a=0$ when $F=0$. This formulation prioritizes adequate forecasts relative to the history-only reference while distinguishing resource use among agents with different levels of predictive performance. For forecasts matching or exceeding the reference score, efficiency differences reflect execution time, token consumption, redundant operations, and repeated errors. The default total score is
$S=0.60F+0.20P+0.15T+0.05E$.
Invalid final forecasts receive zero total and efficiency scores.

\subsection{Baselines}
\label{sec:baselines}
For model-level evaluation, we compare 15 LLMs within the same CastWorld harness on the 250 episodes described in Section~\ref{sec:evaluation-datasets}. For system-level evaluation, Table~\ref{tab:forecasting-main-preview} compares 23 forecasting methods with the CastWorld reference agent powered by DeepSeek-V4-Flash. STAT, DL, and TSFM denote statistical models, deep learning models, and time series foundation models, respectively; the LLM-based category also includes agentic systems. The reference agent autonomously selects and uses CastWorld's contextual resources and tools to construct and refine forecasts. The system-level comparison covers 1,310 episodes across 25 datasets and reports macro-averaged MASE and RMSSE alongside average rank.

\begin{figure*}[t]
  \centering
  \includegraphics[width=\textwidth]{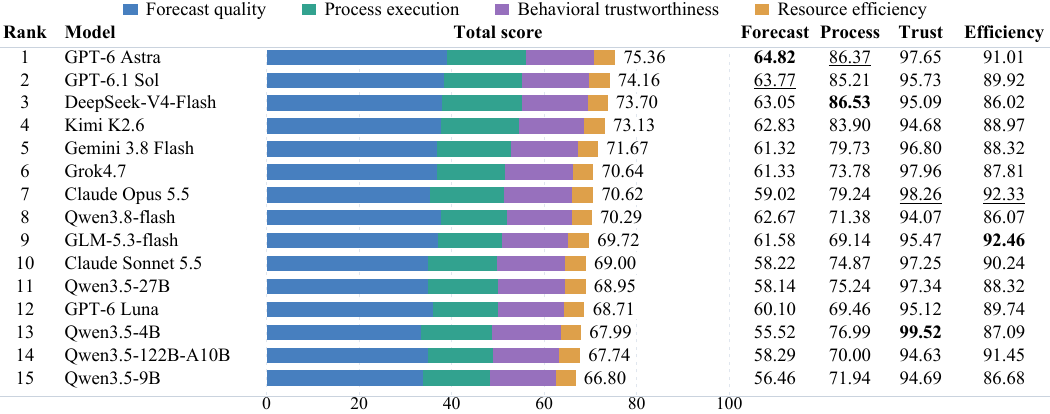}
  \par
  \vspace{-10pt}
  \caption{LLM comparison beyond forecast accuracy on CastWorld. Leaders differ across dimensions. Bars show weighted total-score contributions; columns show unweighted scores. Bold and underlined values mark the best and second-best scores.}
  \Description{Fifteen LLMs ranked by total score, with stacked contributions and four aligned dimension-score columns.}
  \label{fig:llm-leaderboard}
  \vspace{-2pt}
\end{figure*}

\begin{table}[!t]
   \par
   \vspace{-5pt}
  \caption{Forecast comparison across 25 datasets. CastWorld's reference agent achieves the lowest macro-averaged MASE and RMSSE and the best average rank.}
  \label{tab:forecasting-main-preview}
  \centering
 \vspace{-5pt}
  \begingroup
  \setlength{\tabcolsep}{3.75pt}
  \renewcommand{\arraystretch}{0.97}
  \makeatletter
  \newcommand{\cwDash}{%
    \rlap{%
      \kern-\tabcolsep
      \raisebox{\ht\@arstrutbox}[0pt][0pt]{%
        \hbox to\columnwidth{%
          \leaders\hbox{\rule{1.5pt}{.25pt}\kern1.5pt}\hfill}}}%
  }
  \newcommand{\cwEvenType}[1]{%
    \raisebox{\dimexpr-\ht\@arstrutbox/2-\dp\@arstrutbox/2\relax}[0pt][0pt]{#1}%
  }
  \makeatother
  \begin{tabular*}{\columnwidth}{@{\extracolsep{\fill}\hspace{\tabcolsep}}llrrr}
    \toprule
    Type & Method & MASE & RMSSE & Avg. rank \\
    \midrule
    STAT & ARIMAX~\cite{pankratz1991dynamic} & 1.7129 & 1.3703 & 15.45 \\
    \cwDash & DLinear~\cite{zeng2023dlinear} & 1.2532 & 1.0043 & 11.86 \\
    \cwEvenType{DL} & PatchTST~\cite{nie2023patchtst} & 1.1032 & 0.8668 & 10.53 \\
     & iTransformer~\cite{liu2024itransformer} & 1.3708 & 1.0714 & 10.90 \\
     & TimeXer~\cite{wang2024timexer} & 1.1127 & 0.8862 & 9.40 \\
    \cwDash & TimesFM~\cite{das2024timesfm} & 1.2609 & 1.0187 & 13.56 \\
    \cwEvenType{TSFM} & Moirai~\cite{woo2024moirai} & 1.2206 & 0.9939 & 13.70 \\
     & Sundial~\cite{liu2025sundial} & 1.2729 & 1.0099 & 12.97 \\
     & Chronos~\cite{ansari2024chronos} & 0.9910 & 0.7977 & 7.85 \\
    \cwDash & PromptCast~\cite{xue2024promptcast} & 1.0146 & 0.8312 & 13.32 \\
     & Time-LLM~\cite{jin2024timellm} & 1.2475 & 1.0086 & 15.58 \\
     & TokenCast~\cite{tao2026tokencast} & 1.3221 & 1.0611 & 17.88 \\
     & LLMTime~\cite{gruver2023llmtime} & 1.6679 & 1.3108 & 22.00 \\
     & TimeReasoner~\cite{cheng2026timereasoner} & 1.0649 & 0.8595 & 13.16 \\
     & Time-R1~\cite{zhou2025timer1} & 0.9539 & 0.7691 & 12.15 \\
    \makebox[26.64907pt][l]{LLM} & MemCast~\cite{tao2026memcast} & 0.9092 & 0.7400 & \underline{5.78} \\
    based & AlphaCast~\cite{zhang2025alphacast} & 0.9738 & 0.7846 & 8.36 \\
     & TimeCopilot~\cite{garza2025timecopilot} & 1.2886 & 1.0227 & 18.17 \\
     & TimeSeriesScientist~\cite{zhao2025timeseriesscientist} & 1.4223 & 1.1617 & 19.13 \\
     & Cast-R1~\cite{tao2026castr1} & 1.0669 & 0.8674 & 14.28 \\
     & CastFlow~\cite{pan2026castflow} & 1.0826 & 0.8583 & 10.12 \\
     & MoiraiAgent~\cite{salesforce2026moiraiagent} & \underline{0.8963} & \underline{0.7396} & 8.23 \\
     & CastFSR~\cite{tao2026castfsr} & 1.0983 & 0.8876 & 12.56 \\
    \midrule
    Ours & CastWorld agent & \textbf{0.8284} & \textbf{0.6869} & \textbf{3.06} \\
    \bottomrule
  \end{tabular*}
  \endgroup
  \vspace{-15pt}
\end{table}

\subsection{Implementation Details}
\label{sec:implementation-details}

Model-level evaluation uses a default budget of 30 interaction rounds per forecasting episode, with at most four read-only calls per round. The harness provides frozen native forecast candidates from Chronos T5 Base, TimesFM 2.5 200M, Sundial Base 128M, and Moirai 1.1-R Base. Budget and evidence-access studies comprise 2,152 evaluations across eight datasets using DeepSeek-V4-Flash (DeepSeek), Qwen3.5-9B (Qwen), and GPT-6 Astra (Astra). These studies pair the same tasks across conditions within each model. Accuracy-matched behavioral comparisons use a common 48-task panel. MASE, RMSSE, and scores are averaged within datasets and then equally across datasets; first-forecast rounds and forecast-version counts are averaged across episodes. For the system-level comparison, average rank aggregates 50 dataset--metric rankings, assigning average ranks to ties.

Diagnostic curves use completed trajectories. We separately report forecast availability within the ordinary interaction budget, delivery under the declared completion and repair protocol, and supplementary completion runs. Appendices~\ref{sec:diagnostic-protocols} and~\ref{sec:resource-analysis} provide model-specific task counts, completion procedures, and token accounting, including estimates of input and output token usage.

\vspace{-5pt}
\section{Experimental Results}
\label{sec:experiments}

We compare forecasting systems and LLM-driven agents, then examine agent behavior at comparable accuracy, context and tool ablations, and forecasting toolkits. We further analyze context-guided revision, adaptation to interaction budgets, and evidence access during refinement. Appendices~\ref{sec:historical-variation} and~\ref{sec:robustness} report historical task variation, repeatability, and window sensitivity.

\vspace{-5pt}
\subsection{Model- and System-Level Performance}
\label{sec:main-results}

As shown in Table~\ref{tab:forecasting-main-preview}, CastWorld's reference agent achieves the lowest macro-averaged MASE and RMSSE and the best average rank across the 25 datasets. The strongest comparator differs between macro-averaged error and average rank, reflecting the complementary information captured by these summaries. Macro-averaged errors retain the magnitude of scaled prediction errors, whereas average rank summarizes relative standing within individual dataset--metric comparisons. The following comparison evaluates 15 LLMs under a shared forecasting harness and evaluation protocol.

\begin{figure}[t]
  \centering
  \includegraphics[width=\columnwidth]{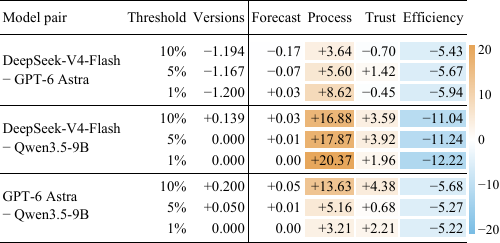}
  \par
  \vspace{-4pt}
  \caption{Comparing agents at matched MASE and RMSSE. Values are first minus second for scores and mean agent-created version counts. Thresholds bound relative error gaps.}
  \Description{Three model pairs matched by symmetric relative gaps in both errors, showing differences in forecast versions and the four evaluation dimensions.}
  \label{fig:quality-matched}
  \vspace{-10pt}
\end{figure}

As shown in Figure~\ref{fig:llm-leaderboard}, no LLM leads across all four dimensions. Under shared tool access, differences in forecast quality highlight the importance of effective tool selection and use. Nearly identical forecast errors can also accompany different process scores, supporting the joint evaluation of predictive performance and forecasting behavior. The resource records in Table~\ref{tab:llm-resources} show that similar tool-call volumes can accompany different numbers of agent-created forecast versions. For example, GPT-6.1 Sol and Claude Sonnet 5.5 average 14.94 and 14.53 tool calls per episode, while creating 1.41 and 1.04 forecast versions, respectively. We next examine these differences in decision processes and resource use after explicitly matching forecast accuracy.

\vspace{-5pt}
\subsection{Agent Behavior at Comparable Accuracy}
\label{sec:matched-results}

To examine decision processes at comparable accuracy, we match valid predictions from the 48 tasks shared by DeepSeek-V4-Flash, Qwen3.5-9B, and GPT-6 Astra. Both MASE and RMSSE must satisfy $\frac{2|u-v|}{u+v}\leq\epsilon$, for $\epsilon\in\{0.10,0.05,0.01\}$, where $u$ and $v$ are the corresponding errors of the two agents on the same task. As shown in Figure~\ref{fig:quality-matched}, matched agents still differ in forecast construction and resource use. At the primary 10\% threshold, DeepSeek scores 16.88 points higher in process execution and 11.04 points lower in efficiency than Qwen. At 1\%, their reported mean counts of agent-created forecast versions are equal, while their process scores differ by 20.37 points.

At the primary 10\% matching threshold, the process-score differences persist after removing final-selection quality and after restricting evaluation to checks applicable to both trajectories. These comparisons keep forecasts and matched tasks fixed, allowing differences in recorded actions to be examined under common evaluation criteria. The common-check decomposition places most of the DeepSeek--Qwen gap in public checking, with a smaller contribution from need--query alignment. These results reveal limitations in agents' effective use of forecasting tools, particularly in checking submitted or related forecasts and aligning tool requests with declared information needs. Appendix~\ref{sec:process-selection-ablation} reports all thresholds and shared-check comparisons.

Tighter matching leaves fewer pairs and changes which tasks qualify, so we use the 10\% matches as the primary descriptive comparison. Expert verification provides a separate check on the rules used to assess recorded behavior. Automated verdicts closely agree with unanimous applicable judgments of actions and numerical consistency. Appendices~\ref{sec:diagnostic-protocols} and~\ref{sec:behavior-check-validation} give the matching counts and per-check agreement.

\begin{figure}[t]
  \centering
  \includegraphics[width=\columnwidth]{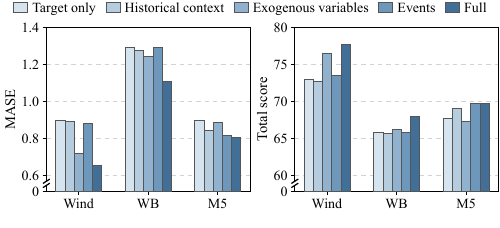}
  \vspace{-23pt}
  \caption{Effect of contextual information on forecasting. All conditions retain target history. Full achieves the lowest MASE on all three tasks.}
  \Description{Two grouped-bar panels compare MASE and total score under five context conditions, preserving target history in every condition.}
  \label{fig:context-ablation}
  \vspace{-10pt}
\end{figure}

\begin{figure}[t]
  \centering
  \includegraphics[width=\columnwidth]{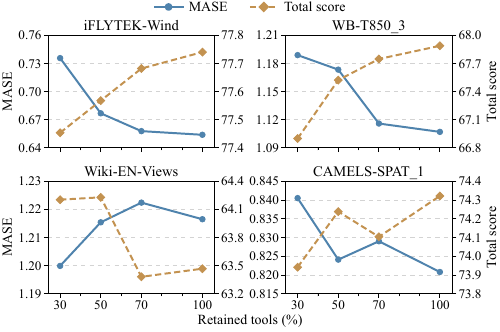}
  \vspace{-0.3in}
  \caption{Effect of tool inventory size on forecasting. Inventories are fixed and nested. MASE uses the left axis and total score the right. Panel ranges differ.}
  \Description{Four dual-axis panels show MASE and total score as the retained tool inventory increases from 30 to 100 percent.}
  \label{fig:tool-inventory}
  \vspace{-0.22in}
\end{figure}

\subsection{Context and Tool Ablations}
\label{sec:context-results}

Context, tool, and state-summary ablations use the reference agent. Historical context restricts exogenous variables and events to the observed interval, while the two channel-specific conditions retain either exogenous variables or events. All conditions preserve target history. As shown in Figure~\ref{fig:context-ablation}, the Full condition achieves the lowest errors across the displayed tasks, outperforming both the target-only and partial-context conditions. Exogenous variables are more helpful for the physical forecasting tasks, whereas events contribute more to the retail task. These task-dependent benefits support including both context channels and enabling agents to select relevant evidence.

\begin{figure*}[t]
  \centering
  \includegraphics[width=\textwidth,height=195.05525pt]{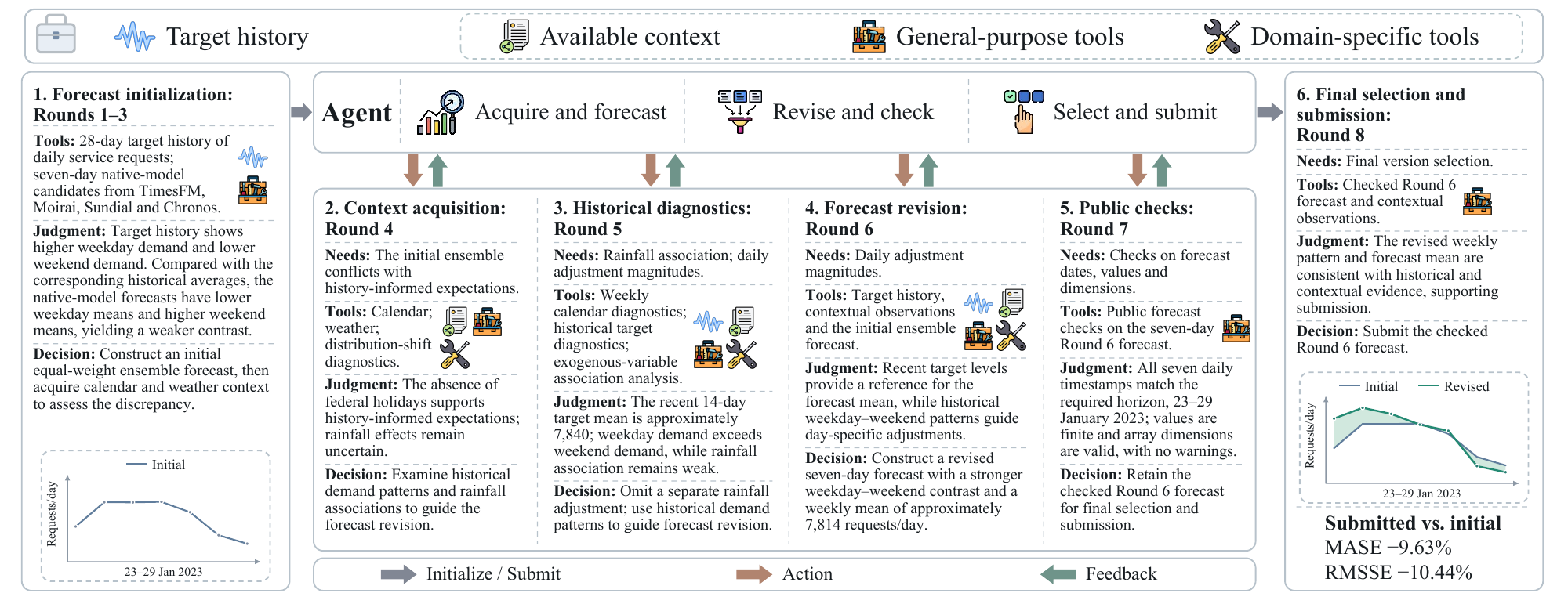}
  \vspace{-10pt}
  \caption{Context-guided revision connects information acquisition to numerical changes on NYC-311. Recorded observations support a revised daily pattern with lower offline errors than the initial ensemble.}
  \Description{A demand-forecasting trajectory uses calendar and weather observations, historical weekly patterns, and level diagnostics to revise a seven-day forecast. Weak rainfall association is not used for a separate adjustment. Offline MASE and RMSSE decrease by 9.63 and 10.44 percent.}
  \label{fig:case-study}
\end{figure*}

\begin{figure}[t]
  \centering
  \includegraphics[width=\columnwidth]{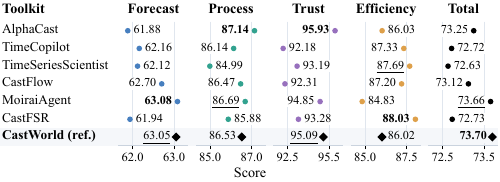}
  \vspace{-10pt}
  \caption{Comparison of forecasting toolkits. Best and second-best scores are bold and underlined, respectively. Black diamonds denote CastWorld's reference toolkit. Axis ranges differ across panels.}
  \Description{Five aligned horizontal dot plots in a single row compare seven forecasting toolkits. Each point is labeled with its score, with the best and second-best scores in each panel bold and underlined, respectively. CastWorld (ref.) denotes the benchmark's reference toolkit, highlighted by black diamonds and a shaded row.}
  \label{fig:toolkit-comparison}
  \vspace{-5pt}
\end{figure}

Fixed nested inventories retain approximately 70\%, 50\%, or 30\% of the tool catalog, preserving essential forecast creation, checking, submission, and all four native forecasters. As shown in Figure~\ref{fig:tool-inventory}, the full inventory achieves the lowest MASE on three of the four displayed datasets. Forecast accuracy improves consistently with inventory size on the wind and temperature tasks, while the Web-audience task benefits from a reduced inventory. These results show the benefits of broader tool availability across multiple tasks, alongside task-dependent responses to inventory size. Total scores can remain nearly unchanged even when raw error increases because they also reflect other aspects of the interaction. Resource-family ablations further show higher MASE and RMSSE on all three tested tasks after removing either general-purpose tools or domain resources, including knowledge. Both resource families thus contribute to the reference agent's forecasting accuracy. Appendix~\ref{sec:tool-analysis} reports these results and the state-summary control.

\subsection{Comparison of Forecasting Toolkits}
\label{sec:toolkit-comparison}

We compare CastWorld's reference toolkit with those of AlphaCast~\cite{zhang2025alphacast}, TimeCopilot~\cite{garza2025timecopilot}, TimeSeriesScientist~\cite{zhao2025timeseriesscientist}, CastFlow~\cite{pan2026castflow}, MoiraiAgent~\cite{salesforce2026moiraiagent}, and CastFSR~\cite{tao2026castfsr} in CastWorld's forecasting environment. All toolkits are evaluated using DeepSeek-V4-Flash on the same 250 episodes under an identical interaction budget and the same four-dimensional scoring protocol. This comparison examines the toolkits' support for context-aware forecasting through both forecast quality and the execution, trustworthiness, and resource efficiency of agent interactions.

As shown in Figure~\ref{fig:toolkit-comparison}, CastWorld's reference toolkit achieves the highest total score (73.70), ranking second in forecast quality (63.05) and behavioral trustworthiness (95.09) and third in process execution (86.53). Its forecast-quality score is close to MoiraiAgent's leading score of 63.08, with higher behavioral trustworthiness and resource efficiency by 0.24 and 1.19 points, respectively. AlphaCast leads in process execution and behavioral trustworthiness, while CastWorld attains a 1.17-point higher forecast-quality score. These comparisons reveal distinct performance profiles across toolkits, including differences in interaction quality and efficiency at similar forecast quality. CastWorld combines competitive forecasting with strong process execution and behavioral trustworthiness, yielding the highest aggregate score under the shared evaluation protocol. This combination supports its use as a foundational toolkit for context-aware forecasting agents.

\subsection{Context-Guided Forecast Revision}
\label{sec:trajectory-diagnostics}

The trajectory in Figure~\ref{fig:case-study} illustrates how contextual observations inform a numerical revision. DeepSeek-V4-Flash forecasts a week of NYC-311 service requests from 28 daily observations. The initial ensemble conflicts with the agent's reading of recent demand, prompting calendar and weather queries followed by historical diagnostics. These observations support a stable demand level with a weekday--weekend contrast, while the rainfall association is weak. The agent reconstructs the weekly pattern without applying a separate rainfall correction.

The submitted forecast raises Monday's value while retaining a weekly mean close to recent demand. It changes the daily pattern rather than uniformly increasing the forecast. Public checks verify dates, values, and dimensions before finalization. Offline evaluation shows that the submitted forecast reduces MASE by 9.63\% and RMSSE by 10.44\% relative to the initial ensemble. Held-out targets remain unavailable during construction. Recorded observations and rationales make the change between the two forecast versions inspectable, linking the agent's interpretation of context to the resulting numerical changes.

\subsection{Adaptation to Interaction Budgets}
\label{sec:budget-results}

\begin{figure}[t]
  \centering
  \includegraphics[width=\columnwidth]{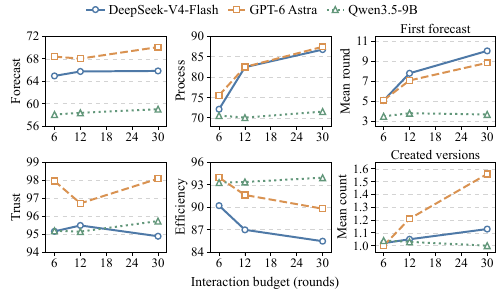}
  \caption{Effects of interaction budgets on forecasting behavior. Across three LLMs, the panels compare evaluation scores, first-forecast timing, and agent-created version counts.}
  \vspace{-0.12in}
  \Description{Six panels show score profiles, mean first-forecast round, and mean agent-created versions under budgets of six, twelve, and thirty rounds.}
  \label{fig:budget-adaptation}
\end{figure}

We vary the interaction budget over 6, 12, and 30 rounds on a fixed task cohort for each model. Tighter budgets reduce the availability of valid forecasts within ordinary rounds, while all three models reach full availability on their respective cohorts at 30 rounds. Appendix~\ref{sec:diagnostic-protocols} reports these counts and distinguishes this availability from delivery after declared repair. DeepSeek's MASE is 0.0389 higher with 6 rounds than with 30 rounds, with a 95\% paired block-bootstrap confidence interval of $[0.0094, 0.0706]$.

As shown in Figure~\ref{fig:budget-adaptation}, the models use additional rounds differently. As the budget increases from 6 to 30 rounds, Astra delays its first forecast and increases its mean agent-created version count from 1.00 to 1.56, whereas Qwen largely retains its early, near-single-version strategy. DeepSeek also postpones initial construction, while its mean version count increases from 1.02 to 1.13. Its process score improves, while raw-error gains beyond the intermediate budget remain small. A longer trajectory can therefore reflect more work before forecasting rather than repeated numerical refinement. Extended interaction also lowers efficiency for the agents that substantially increase their resource use.

\subsection{Evidence Access During Refinement}
\label{sec:workflow-analysis}

\begin{figure}[t]
  \centering
  \includegraphics[width=\columnwidth]{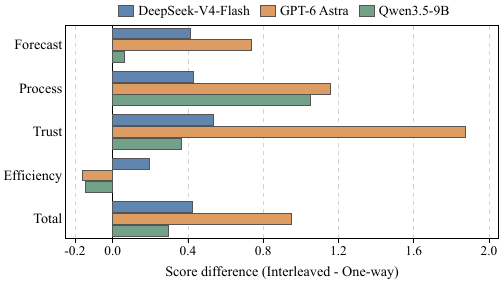}
  \caption{Effect of continued information access. Bars show Interleaved minus One-way. Both are autonomous. One-way blocks external acquisition after the first agent version.}
  \Description{One grouped horizontal bar chart compares three models on a shared difference axis, showing interleaved minus one-way scores in the four dimensions and total score.}
  \label{fig:workflow-control}
  \vspace{-0.14in}
\end{figure}

To test the effect of continued information access during refinement, the workflow control varies whether new external information remains accessible after the first agent-created forecast is saved. Both conditions allow 30 rounds and autonomous decisions. One-way freezes new external acquisition at that point. Native candidates visible before adoption do not trigger the freeze. Existing observations, same-history native predictions, revision, checks, and final selection remain available. Interleaved interaction retains access to new information during forecast refinement. As shown in Figure~\ref{fig:workflow-control}, continued information access increases task-aware forecast scores and total scores in all three model-specific comparisons. Qwen's forecast-score improvement accompanies an increase in MASE. For DeepSeek, the paired One-way-minus-Interleaved MASE difference is 0.0224, with a 95\% block-bootstrap confidence interval of $[-0.0043, 0.0517]$. The interval uses 2,000 temporal-block resamples with fixed task pairs and equal dataset weights.

Across 250 episodes per model, 12 of the 15 models make information-tool calls after their first forecast in fewer than one quarter of episodes, indicating limited continued context acquisition during iterative forecasting. In the common 48-task panel, Astra and DeepSeek numerically revise their forecasts in 19 and 9 episodes, respectively. In most of these episodes, the agents improve on the first forecast and submit the best generated version under offline $F$. In 4 Astra and 3 DeepSeek episodes, they instead submit lower-scoring revisions while the first version remains best. Qwen produces one agent-created version per episode. Across the full 15-model evaluation, most models fail to select their best generated forecast under offline $F$ in some episodes, revealing limitations in forecast self-evaluation and final selection. Appendix~\ref{sec:revision-results} provides complementary error-based version comparisons.

\section{Conclusion}
\label{sec:conclusion}
In this paper, we introduced CastWorld, a unified benchmark for context-aware time series forecasting agents. CastWorld integrates datasets with multidimensional real-world context, a standardized and extensible forecasting harness, and systematic model- and system-level evaluation. Its state-adaptive loop allows intermediate forecasts to guide evidence acquisition and new evidence to inform forecast refinement, while decision trajectories and forecast-version histories enable evaluation of the complete forecasting process. Experiments reveal limitations in continuous context acquisition, effective forecasting tool selection and use, and self-evaluation for identifying and submitting the best generated forecast version. These findings identify key challenges in translating contextual resources and tool capabilities into reliable predictions. CastWorld provides a shared foundation for studying these capabilities and advancing autonomous forecasting in real-world environments, supporting Temporal World Model research on context-aware prediction of system dynamics.

\bibliographystyle{ACM-Reference-Format}
\bibliography{references}

@inproceedings{godahewa2021monash,
  author = {Godahewa, Rakshitha and Bergmeir, Christoph and Webb, Geoffrey I. and Hyndman, Rob J. and Montero-Manso, Pablo},
  title = {Monash Time Series Forecasting Archive},
  year = {2021},
  booktitle = {Proceedings of the Neural Information Processing Systems Track on Datasets and Benchmarks},
  volume = {1},
}

@inproceedings{liu2024agentbench,
  author = {Liu, Xiao and Yu, Hao and Zhang, Hanchen and Xu, Yifan and Lei, Xuanyu and Lai, Hanyu and Gu, Yu and Ding, Hangliang and Men, Kaiwen and Yang, Kejuan and Zhang, Shudan and Deng, Xiang and Zeng, Aohan and Du, Zhengxiao and Zhang, Chenhui and Shen, Sheng and Zhang, Tianjun and Su, Yu and Sun, Huan and Huang, Minlie and Dong, Yuxiao and Tang, Jie},
  title = {{AgentBench}: Evaluating {LLM}s as Agents},
  year = {2024},
  booktitle = {International Conference on Learning Representations},
  pages = {52989--53046},
}

@article{ma2024agentboard,
  author = {Ma, Chang and Zhang, Junlei and Zhu, Zhihao and Yang, Cheng and Yang, Yujiu and Jin, Yaohui and Lan, Zhenzhong and Kong, Lingpeng and He, Junxian},
  title = {{AgentBoard}: An Analytical Evaluation Board of Multi-turn {LLM} Agents},
  year = {2024},
  journal = {Advances in Neural Information Processing Systems},
  volume = {37},
  pages = {74325--74362},
}

@inproceedings{zhou2024webarena,
  author = {Zhou, Shuyan and Xu, Frank F and Zhu, Hao and Zhou, Xuhui and Lo, Robert and Sridhar, Abishek and Cheng, Xianyi and Ou, Tianyue and Bisk, Yonatan and Fried, Daniel and Alon, Uri and Neubig, Graham},
  title = {{WebArena}: A Realistic Web Environment for Building Autonomous Agents},
  year = {2024},
  booktitle = {International Conference on Learning Representations},
  pages = {15585--15606},
}

@inproceedings{yao2025taubench,
  author = {Yao, Shunyu and Shinn, Noah and Razavi, Pedram and Narasimhan, Karthik},
  title = {{$\tau$-bench}: A Benchmark for Tool-Agent-User Interaction in Real-World Domains},
  year = {2025},
  booktitle = {International Conference on Learning Representations},
  pages = {9965--10017},
}

@inproceedings{kapoor2026hal,
  author = {Kapoor, Sayash and Stroebl, Benedikt and Kirgis, Peter and Nadgir, Nitya and Siegel, Zachary and Wei, Boyi and Xue, Tianci and Chen, Ziru and Chen, Felix and Utpala, Saiteja and Ndzomga, Franck and Oruganty, Dheeraj and Luskin, Sophie and Liu, Kangheng and Yu, Botao and Arora, Amit and Hahm, Dongyoon and Trivedi, Harsh and Sun, Huan and Lee, Juyong and Jin, Tengjun and Mai, Yifan and Zhou, Yifei and Zhu, Yuxuan and Bommasani, Rishi and Kang, Daniel and Song, Dawn and Henderson, Peter and Su, Yu and Liang, Percy and Narayanan, Arvind},
  title = {Holistic Agent Leaderboard: The Missing Infrastructure for {AI} Agent Evaluation},
  year = {2026},
  booktitle = {International Conference on Learning Representations},
  pages = {98778--98849},
}

@article{qiu2024tfb,
  author = {Qiu, Xiangfei and Hu, Jilin and Zhou, Lekui and Wu, Xingjian and Du, Junyang and Zhang, Buang and Guo, Chenjuan and Zhou, Aoying and Jensen, Christian S. and Sheng, Zhenli and Yang, Bin},
  title = {{TFB}: Towards Comprehensive and Fair Benchmarking of Time Series Forecasting Methods},
  year = {2024},
  journal = {Proceedings of the VLDB Endowment},
  volume = {17},
  number = {9},
  pages = {2363--2377},
}

@inproceedings{williams2025cik,
  author = {Williams, Andrew Robert and Ashok, Arjun and Marcotte, \'{E}tienne and Zantedeschi, Valentina and Subramanian, Jithendaraa and Riachi, Roland and Requeima, James and Lacoste, Alexandre and Rish, Irina and Chapados, Nicolas and Drouin, Alexandre},
  title = {Context is Key: A Benchmark for Forecasting with Essential Textual Information},
  year = {2025},
  booktitle = {Proceedings of the 42nd International Conference on Machine Learning},
  pages = {66887--66944},
}

@article{tang2026drcik,
  author = {Tang, Yihong and Williams, Andrew Robert and Ashok, Arjun and Zheng, Vincent Zhihao and Gupta, Shubham and Sun, Lijun and Drouin, Alexandre and Laradji, Issam H. and Marcotte, \'{E}tienne and Zantedeschi, Valentina},
  title = {{Dr-CiK}: A Benchmark testing Deep Research for Context-Aided Forecasting},
  year = {2026},
  journal = {arXiv preprint arXiv:2605.27904},
}

@article{tao2026castfsr,
  author = {Tao, Xiaoyu and Cheng, Mingyue and Pan, Bokai and Jiang, Chuang and Zhang, Huanjian and Gao, Tian and Liu, Yaguo and Liu, Qi and Chen, Enhong},
  title = {{CastFSR}: A Fast--Slow--Reflect Agentic Reasoning Framework for Context-Aware Time Series Forecasting},
  year = {2026},
  journal = {arXiv preprint arXiv:2608.03031v2},
}

@inproceedings{lu2025toolsandbox,
  author = {Lu, Jiarui and Holleis, Thomas and Zhang, Yizhe and Aumayer, Bernhard and Nan, Feng and Bai, Haoping and Ma, Shuang and Ma, Shen and Li, Mengyu and Yin, Guoli and Wang, Zirui and Pang, Ruoming},
  title = {{ToolSandbox}: A Stateful, Conversational, Interactive Evaluation Benchmark for {LLM} Tool Use Capabilities},
  year = {2025},
  booktitle = {Findings of the Association for Computational Linguistics: NAACL 2025},
  pages = {1160--1183},
}

@inproceedings{chen2026trace,
  author = {Chen, Yanyu and Jiang, Jiyue and Liu, Jiahong and Zhang, Yifei and Guo, Xiao and King, Irwin},
  title = {{TRACE}: Trajectory-Aware Comprehensive Evaluation for Deep Research Agents},
  year = {2026},
  booktitle = {Proceedings of the ACM Web Conference 2026},
  pages = {2524--2534},
}

@article{liu2024timemmd,
  author = {Liu, Haoxin and Xu, Shangqing and Zhao, Zhiyuan and Kong, Lingkai and Kamarthi, Harshavardhan and Sasanur, Aditya B. and Sharma, Megha and Cui, Jiaming and Wen, Qingsong and Zhang, Chao and Prakash, B. Aditya},
  title = {{Time-MMD}: Multi-Domain Multimodal Dataset for Time Series Analysis},
  year = {2024},
  journal = {Advances in Neural Information Processing Systems},
  volume = {37},
  pages = {77888--77933},
}

@article{liu2026giftctx,
  author = {Liu, Xu and Yan, Hanshu and Aksu, Taha and Pham, Quang and Qiao, Zhongzheng and Liew, Jun Hao and Savarese, Silvio and Sahoo, Doyen and Li, Junnan},
  title = {{GIFT-Eval-CTX}: A Benchmark for Contextual Forecasting Model Evaluation},
  year = {2026},
  journal = {arXiv preprint arXiv:2610.17794},
}

@article{nagashima2026fwbench,
  author = {Nagashima, Shunya},
  title = {Forecast Workflow Bench: Evaluating Language-Model Decisions with Budgeted Forecast Tools},
  year = {2026},
  journal = {arXiv preprint arXiv:2609.27385},
}

@article{wang2025paperarena,
  author = {Wang, Daoyu and Cheng, Mingyue and Yu, Shuo and Liu, Zirui and Guo, Ze and Li, Xin and Liu, Qi},
  title = {{PaperArena}: An Evaluation Benchmark for Tool-Augmented Agentic Reasoning on Scientific Literature},
  year = {2025},
  journal = {arXiv preprint arXiv:2510.10909},
}

@article{zhang2024probts,
  author = {Zhang, Jiawen and Wen, Xumeng and Zhang, Zhenwei and Zheng, Shun and Li, Jia and Bian, Jiang},
  title = {{ProbTS}: Benchmarking Point and Distributional Forecasting across Diverse Prediction Horizons},
  year = {2024},
  journal = {Advances in Neural Information Processing Systems},
  volume = {37},
  pages = {48045--48082},
}

@article{zhu2025abc,
  author = {Zhu, Yuxuan and Jin, Tengjun and Pruksachatkun, Yada and Zhang, Andy and Liu, Shu and Cui, Sasha and Kapoor, Sayash and Longpre, Shayne and Meng, Kevin and Weiss, Rebecca and Barez, Fazl and Gupta, Rahul and Dhamala, Jwala and Merizian, Jacob and Giulianelli, Mario and Coppock, Harry and Ududec, Cozmin and Kellermann, Antony and Sekhon, Jasjeet and Steinhardt, Jacob and Schwettmann, Sarah and Narayanan, Arvind and Zaharia, Matei A and Stoica, Ion and Liang, Percy and Kang, Daniel},
  title = {Establishing Best Practices in Building Rigorous Agentic Benchmarks},
  year = {2025},
  journal = {Advances in Neural Information Processing Systems},
  volume = {38},
}

@inproceedings{li2025tsfmbench,
  author = {Li, Zhe and Qiu, Xiangfei and Chen, Peng and Wang, Yihang and Cheng, Hanyin and Shu, Yang and Hu, Jilin and Guo, Chenjuan and Zhou, Aoying and Jensen, Christian S. and Yang, Bin},
  title = {{TSFM-Bench}: A Comprehensive and Unified Benchmark of Foundation Models for Time Series Forecasting},
  year = {2025},
  booktitle = {Proceedings of the 31st ACM SIGKDD Conference on Knowledge Discovery and Data Mining V.2},
  pages = {5595--5606},
}

@article{gruver2023llmtime,
  author = {Gruver, Nate and Finzi, Marc and Qiu, Shikai and Wilson, Andrew G},
  title = {Large Language Models Are Zero-Shot Time Series Forecasters},
  year = {2023},
  journal = {Advances in Neural Information Processing Systems},
  volume = {36},
  pages = {19622--19635},
}

@article{requeima2024llmprocesses,
  author = {Requeima, James and Bronskill, John and Choi, Dami and Turner, Richard E. and Duvenaud, David},
  title = {{LLM} Processes: Numerical Predictive Distributions Conditioned on Natural Language},
  year = {2024},
  journal = {Advances in Neural Information Processing Systems},
  volume = {37},
  pages = {109609--109671},
}

@inproceedings{cheng2026timereasoner,
  author = {Cheng, Mingyue and Wang, Jiahao and Wang, Daoyu and Tao, Xiaoyu and Liu, Qi and Chen, Enhong},
  title = {Can Slow-Thinking {LLM}s Reason Over Time? Empirical Studies in Time Series Forecasting},
  year = {2026},
  booktitle = {Proceedings of the Nineteenth ACM International Conference on Web Search and Data Mining},
  pages = {99--110},
}

@article{tao2026castr1,
  author = {Tao, Xiaoyu and Cheng, Mingyue and Jiang, Chuang and Gao, Tian and Zhang, Huanjian and Liu, Yaguo},
  title = {{Cast-R1}: Learning Tool-Augmented Sequential Decision Policies for Time Series Forecasting},
  year = {2026},
  journal = {arXiv preprint arXiv:2602.13802v1},
}

@inproceedings{jalori2025flairrts,
  author = {Jalori, Gunjan and Verma, Preetika and Arik, Sercan O},
  title = {{FLAIRR-TS} - Forecasting {LLM}-Agents with Iterative Refinement and Retrieval for Time Series},
  year = {2025},
  booktitle = {Findings of the Association for Computational Linguistics: EMNLP 2025},
  pages = {15427--15437},
}

@inproceedings{weng2026temporalbench,
  author = {Weng, Muyan and Cao, Defu and Yang, Wei and Sharma, Yashaswi and Liu, Yan},
  title = {{TemporalBench}: A Benchmark for Evaluating {LLM}-Based Agents on Contextual and Event-Informed Time Series Tasks},
  year = {2026},
  booktitle = {Proceedings of the 32nd ACM SIGKDD Conference on Knowledge Discovery and Data Mining V.2},
  pages = {9997--10008},
}

@inproceedings{ahamed2026tfrbench,
  author = {Ahamed, Md Atik and Parmar, Mihir and Goyal, Palash and Song, Yiwen and Le, Long T. and Cheng, Qiang and Li, Chun-Liang and Palangi, Hamid and Yoon, Jinsung and Pfister, Tomas},
  title = {{TFRBench}: A Reasoning Benchmark for Evaluating Forecasting Systems},
  year = {2026},
  booktitle = {Proceedings of the 43rd International Conference on Machine Learning},
  pages = {1169--1223},
}

@article{cai2025timeseriesgym,
  author = {Cai, Yifu and Li, Xinyu and Goswami, Mononito and Wili{\'n}ski, Micha{\l} and Welter, Gus and Dubrawski, Artur},
  title = {{TimeSeriesGym}: A Scalable Benchmark for (Time Series) Machine Learning Engineering Agents},
  year = {2025},
  journal = {arXiv preprint arXiv:2505.13291},
}

@article{kong2026timesagemt,
  author = {Kong, Yaxuan and Yao, Qingren and Nie, Yuqi and Li, Yichen and Shao, Yilei and Zohren, Stefan and Vettoruzzo, Anna and Vanschoren, Joaquin and Jin, Ming and Wen, Qingsong},
  title = {{TimeSage-MT}: A Multi-Turn Benchmark for Evaluating Agentic Time Series Reasoning},
  year = {2026},
  journal = {arXiv preprint arXiv:2606.01498},
}

@inproceedings{yao2026timesageev,
  author = {Yao, Qingren and Kong, Yaxuan and Nie, Yuqi and Li, Yichen and Zohren, Stefan and Vettoruzzo, Anna and Wen, Qingsong and Jin, Ming and Vanschoren, Joaquin},
  title = {{TimeSage-EV}: A Live Benchmark for Agentic Time Series Analysis in Evolving Environments},
  year = {2026},
  booktitle = {Findings of the Association for Computational Linguistics: EMNLP 2026},
  pages = {23542--23567},
}

@article{ansari2024chronos,
  author = {Ansari, Abdul Fatir and Stella, Lorenzo and Turkmen, Caner and Zhang, Xiyuan and Mercado, Pedro and Shen, Huibin and Shchur, Oleksandr and Rangapuram, Syama Sundar and Pineda Arango, Sebastian and Kapoor, Shubham and Zschiegner, Jasper and Maddix, Danielle C. and Wang, Hao and Mahoney, Michael W. and Torkkola, Kari and Wilson, Andrew Gordon and Bohlke-Schneider, Michael and Wang, Yuyang},
  title = {{Chronos}: Learning the Language of Time Series},
  year = {2024},
  journal = {Transactions on Machine Learning Research},
}

@inproceedings{das2024timesfm,
  author = {Das, Abhimanyu and Kong, Weihao and Sen, Rajat and Zhou, Yichen},
  title = {A Decoder-Only Foundation Model for Time-Series Forecasting},
  year = {2024},
  booktitle = {Proceedings of the 41st International Conference on Machine Learning},
  pages = {10148--10167},
}

@inproceedings{liu2025sundial,
  author = {Liu, Yong and Qin, Guo and Shi, Zhiyuan and Chen, Zhi and Yang, Caiyin and Huang, Xiangdong and Wang, Jianmin and Long, Mingsheng},
  title = {{Sundial}: A Family of Highly Capable Time Series Foundation Models},
  year = {2025},
  booktitle = {Proceedings of the 42nd International Conference on Machine Learning},
  pages = {39295--39317},
}

@inproceedings{woo2024moirai,
  author = {Woo, Gerald and Liu, Chenghao and Kumar, Akshat and Xiong, Caiming and Savarese, Silvio and Sahoo, Doyen},
  title = {Unified Training of Universal Time Series Forecasting Transformers},
  year = {2024},
  booktitle = {Proceedings of the 41st International Conference on Machine Learning},
  pages = {53140--53164},
}

@article{xue2024promptcast,
  author = {Xue, Hao and Salim, Flora D.},
  title = {{PromptCast}: A New Prompt-Based Learning Paradigm for Time Series Forecasting},
  year = {2024},
  journal = {IEEE Transactions on Knowledge and Data Engineering},
  volume = {36},
  number = {11},
  pages = {6851--6864},
}

@inproceedings{jin2024timellm,
  author = {Jin, Ming and Wang, Shiyu and Ma, Lintao and Chu, Zhixuan and Zhang, James Y. and Shi, Xiaoming and Chen, Pin-Yu and Liang, Yuxuan and Li, Yuan-Fang and Pan, Shirui and Wen, Qingsong},
  title = {{Time-LLM}: Time Series Forecasting by Reprogramming Large Language Models},
  year = {2024},
  booktitle = {International Conference on Learning Representations},
  pages = {23857--23880},
}

@inproceedings{huang2025exollm,
  author = {Huang, Qihe and Zhou, Zhengyang and Yang, Kuo and Wang, Yang},
  title = {Exploiting Language Power for Time Series Forecasting with Exogenous Variables},
  year = {2025},
  booktitle = {Proceedings of the ACM on Web Conference 2025},
  pages = {4043--4052},
}

@inproceedings{ma2026quizsf,
  author = {Ma, Shichao and Zhou, Zhengyang and Huang, Qihe and Wang, Binwu and Wang, Yang},
  title = {{QuiZSF}: A Retrieval-Augmented Framework for Zero-Shot Time Series Forecasting},
  year = {2026},
  booktitle = {Proceedings of the ACM Web Conference 2026},
  pages = {1947--1958},
}

@article{ye2026tsreasoner,
  author = {Wen Ye and Wei Yang and Defu Cao and Yizhou Zhang and Lumingyuan Tang and Jie Cai and Yan Liu},
  title = {{TS}-Reasoner: Domain-Oriented Time Series Inference Agents for Reasoning and Automated Analysis},
  year = {2026},
  journal = {Transactions on Machine Learning Research},
}

@article{ye2025tsaia,
  author = {Ye, Wen and Liu, Jinbo and Cao, Defu and Yang, Wei and Liu, Yan},
  title = {When {LLM} Meets Time Series: Can {LLM}s Perform Multi-Step Time Series Reasoning and Inference},
  year = {2025},
  journal = {arXiv preprint arXiv:2509.01822},
}

@article{zeng2023dlinear,
  author = {Ailing Zeng and Muxi Chen and Lei Zhang and Qiang Xu},
  title = {Are Transformers Effective for Time Series Forecasting?},
  year = {2023},
  journal = {Proceedings of the AAAI Conference on Artificial Intelligence},
  volume = {37},
  number = {9},
  pages = {11121--11128},
}

@inproceedings{nie2023patchtst,
  author = {Yuqi Nie and Nam H. Nguyen and Phanwadee Sinthong and Jayant Kalagnanam},
  title = {A Time Series is Worth 64 Words: Long-term Forecasting with Transformers},
  year = {2023},
  booktitle = {International Conference on Learning Representations},
}

@inproceedings{liu2024itransformer,
  author = {Yong Liu and Tengge Hu and Haoran Zhang and Haixu Wu and Shiyu Wang and Lintao Ma and Mingsheng Long},
  title = {{iTransformer}: Inverted Transformers Are Effective for Time Series Forecasting},
  year = {2024},
  booktitle = {International Conference on Learning Representations},
  pages = {11116--11140},
}

@article{wang2024timexer,
  author = {Yuxuan Wang and Haixu Wu and Jiaxiang Dong and Guo Qin and Haoran Zhang and Yong Liu and Yunzhong Qiu and Jianmin Wang and Mingsheng Long},
  title = {{TimeXer}: Empowering Transformers for Time Series Forecasting with Exogenous Variables},
  year = {2024},
  journal = {Advances in Neural Information Processing Systems},
  volume = {37},
  pages = {469--498},
}

@book{pankratz1991dynamic,
  author = {Alan Pankratz},
  title = {Forecasting with Dynamic Regression Models},
  year = {1991},
  publisher = {Wiley},
}

@inproceedings{tao2026tokencast,
  author = {Xiaoyu Tao and Shilong Zhang and Mingyue Cheng and Daoyu Wang and Tingyue Pan and Bokai Pan and Changqing Zhang and Shijin Wang},
  title = {From Values to Tokens: An {LLM}-Driven Framework for Context-Aware Time Series Forecasting via Symbolic Discretization},
  year = {2026},
  booktitle = {Proceedings of the Thirty-Fifth International Joint Conference on Artificial Intelligence},
  pages = {3043--3051},
}

@inproceedings{zhou2025timer1,
  author = {Yitong Zhou and Yucong Luo and Mingyue Cheng and Qi Liu and Jiahao Wang and Daoyu Wang and Enhong Chen},
  title = {Time Series Forecasting via Reasoning: A Slow-Thinking Approach with Reinforcement Fine-Tuned {LLMs}},
  year = {2026},
  booktitle = {Proceedings of the 35th ACM International Conference on Information and Knowledge Management},
  pages = {6667--6678},
}

@inproceedings{tao2026memcast,
  author = {Xiaoyu Tao and Mingyue Cheng and Ze Guo and Shuo Yu and Yaguo Liu and Qi Liu and Shijin Wang},
  title = {{MemCast}: Memory-Driven Time Series Forecasting with Experience-Conditioned Reasoning},
  year = {2026},
  booktitle = {Proceedings of the 43rd International Conference on Machine Learning},
  pages = {119553--119569},
}

@inproceedings{zhang2025alphacast,
  author = {Xiaohan Zhang and Tian Gao and Mingyue Cheng and Bokai Pan and Ze Guo and Yaguo Liu and Xiaoyu Tao and Qi Liu},
  title = {{AlphaCast}: A Human Wisdom-{LLM} Intelligence Co-Reasoning Framework for Time Series Forecasting},
  year = {2026},
  booktitle = {Proceedings of the 35th ACM International Conference on Information and Knowledge Management},
  pages = {6322--6333},
}

@article{garza2025timecopilot,
  author = {Azul Garza and Ren{\'e}e Rosillo},
  title = {\unskip {TimeCopilot}},
  year = {2025},
  journal = {\unskip arXiv preprint arXiv:\allowbreak 2509.00616v3},
}

@article{zhao2025timeseriesscientist,
  author = {Haokun Zhao and Xiang Zhang and Jiaqi Wei and Yiwei Xu and Yuting He and Siqi Sun and Chenyu You},
  title = {{TimeSeriesScientist}: A General-Purpose {AI} Agent for Time Series Analysis},
  year = {2025},
  journal = {arXiv preprint arXiv:2510.01538v2},
}

@article{pan2026castflow,
  author = {Bokai Pan and Mingyue Cheng and Xiaoyu Tao and Zhiding Liu and Shuo Yu and Yuchong Wu and Qi Liu and Defu Lian and Enhong Chen},
  title = {{CastFlow}: Bridging Evidence Acquisition and Accurate Prediction in Agentic Time Series Forecasting},
  year = {2026},
  journal = {arXiv preprint arXiv:2604.27840v3},
}

@online{salesforce2026moiraiagent,
  author = {Hanshu Yan and Hong-Quang Pham and Ibrahim Taha Aksu and Jun Hao Liew and Caiming Xiong and Silvio Savarese and Doyen Sahoo and Junnan Li},
  title = {{MoiraiAgent}: An Agentic Framework for Context-Aware Time-Series Forecasting},
  year = {2026},
  organization = {Salesforce},
  url = {https://www.salesforce.com/blog/moiraiagent/},
  lastaccessed = {November 11, 2026},
}

@article{cheng2026atsf,
  author = {Cheng, Mingyue and Tao, Xiaoyu and Liu, Qi and Guo, Ze and Chen, Enhong},
  title = {Position: Beyond Model-Centric Prediction -- Agentic Time Series Forecasting},
  year = {2026},
  journal = {arXiv preprint arXiv:2602.01776v4},
}

@article{cheng2026tsfsurvey,
  author = {Cheng, Mingyue and Liu, Zhiding and Tao, Xiaoyu and Liu, Qi and Zhang, Jintao and Pan, Tingyue and Zhang, Shilong and He, Panjing and Zhang, Xiaohan and Wang, Daoyu and Wang, Jiahao and Chen, Enhong},
  title = {A Comprehensive Survey of Time Series Forecasting: Concepts, Challenges, and Future Directions},
  year = {2026},
  journal = {IEEE Transactions on Knowledge and Data Engineering},
  volume = {38},
  number = {12},
  pages = {7853--7872},
}

@article{liu2026world2ts,
  author = {Liu, Zhiding and Yang, Jiqian and Cheng, Mingyue and Li, Zhi and Chen, Enhong},
  title = {{World2TS}: Real-World Event Context for Time-Series Forecasting},
  year = {2026},
  journal = {arXiv preprint arXiv:2610.14861},
}

\appendix
\vspace{-15pt}
\section{Appendix}
\label{sec:appendix}

\subsection{Dataset Details}
\label{sec:dataset-resources}

\begin{table*}[t]
  \caption{Dataset statistics and contextual resources across eight domains.}
  \vspace{-8pt}
  \label{tab:dataset-resources}
  \centering
  \begingroup
  \fontsize{8}{9.6}\selectfont
  \renewcommand{\arraystretch}{1.13}
  \setlength{\tabcolsep}{3.2pt}
  \begin{tabular*}{\textwidth}{@{\extracolsep{\fill}\hspace{\tabcolsep}}l*{9}{c}}
    \toprule
    & \multicolumn{3}{c}{\textbf{Forecasting tasks}} & \multicolumn{4}{c}{\textbf{Contextual resources}} & \multicolumn{2}{c}{\textbf{Field access}} \\
    \cmidrule(lr){2-4}\cmidrule(lr){5-8}\cmidrule(lr){9-10}
    Dataset & Interval & $(L,H)$ & Episodes & Fields & Event fields & Knowledge & Cases & History only & Forecast window \\
    \midrule
    \rowcolor[gray]{0.95}
    \multicolumn{10}{l}{\textit{Energy}} \\
    \hspace{3pt}iFLYTEK-PV & 15 min & $(96,96)$ & 50 & 53 & 4 & 65 & 109 & 0 & 53 \\
    \hspace{3pt}iFLYTEK-Wind & 15 min & $(96,96)$ & 50 & 54 & 3 & 66 & 109 & 0 & 54 \\
    \hspace{3pt}CN-GD-Load & 1 h & $(168,24)$ & 50 & 41 & 5 & 53 & 104 & 2 & 39 \\
    \hspace{3pt}OPSD-DELU-Price & 1 h & $(96,24)$ & 65 & 91 & 21 & 103 & 99 & 10 & 81 \\
    \addlinespace[2pt]
    \rowcolor[gray]{0.95}
    \multicolumn{10}{l}{\textit{Natural science}} \\
    \hspace{3pt}WB-T850\_1 & 6 h & $(72,24)$ & 65 & 25 & 1 & 37 & 123 & 0 & 25 \\
    \hspace{3pt}WB-T850\_2 & 6 h & $(72,24)$ & 65 & 25 & 1 & 37 & 123 & 0 & 25 \\
    \hspace{3pt}WB-T850\_3 & 6 h & $(72,24)$ & 65 & 31 & 1 & 43 & 126 & 0 & 31 \\
    \hspace{3pt}SILSO-Sunspot & 1 d & $(30,30)$ & 45 & 23 & 0 & 35 & 82 & 19 & 4 \\
    \addlinespace[2pt]
    \rowcolor[gray]{0.95}
    \multicolumn{10}{l}{\textit{Economics and finance}} \\
    \hspace{3pt}US-RPI & 1 mo & $(48,12)$ & 50 & 142 & 12 & 154 & 57 & 127 & 15 \\
    \hspace{3pt}US-DFF & 1 d & $(30,30)$ & 60 & 45 & 4 & 57 & 98 & 27 & 18 \\
    \hspace{3pt}CN-Port-Export & 1 d & $(30,30)$ & 50 & 30 & 5 & 42 & 43 & 15 & 15 \\
    \addlinespace[2pt]
    \rowcolor[gray]{0.95}
    \multicolumn{10}{l}{\textit{Public services}} \\
    \hspace{3pt}CapitalBikeshare & 1 h & $(168,24)$ & 50 & 25 & 5 & 37 & 117 & 0 & 25 \\
    \hspace{3pt}NYC-311 & 1 d & $(28,7)$ & 50 & 43 & 21 & 55 & 112 & 7 & 36 \\
    \hspace{3pt}US-ICSA & 1 wk & $(52,13)$ & 65 & 100 & 24 & 112 & 19 & 70 & 30 \\
    \hspace{3pt}NYC-EMS & 1 d & $(28,7)$ & 50 & 97 & 22 & 109 & 116 & 35 & 62 \\
    \hspace{3pt}CN-Flu-PosRate & 1 wk & $(52,13)$ & 45 & 32 & 8 & 44 & 85 & 14 & 18 \\
    \addlinespace[2pt]
    \rowcolor[gray]{0.95}
    \multicolumn{10}{l}{\textit{Retail}} \\
    \hspace{3pt}M5-TX3-FOODS & 1 d & $(35,28)$ & 35 & 31 & 15 & 43 & 18 & 9 & 22 \\
    \hspace{3pt}RetailII-NetQty & 1 h & $(168,24)$ & 50 & 33 & 9 & 45 & 91 & 15 & 18 \\
    \addlinespace[2pt]
    \rowcolor[gray]{0.95}
    \multicolumn{10}{l}{\textit{Industrial operations}} \\
    \hspace{3pt}IndustryHeat-ActiveUnits & 1 d & $(30,30)$ & 50 & 65 & 23 & 77 & 35 & 21 & 44 \\
    \hspace{3pt}MotorOil-GR-TI824 & 15 min & $(96,96)$ & 50 & 24 & 0 & 36 & 111 & 24 & 0 \\
    \addlinespace[2pt]
    \rowcolor[gray]{0.95}
    \multicolumn{10}{l}{\textit{Web services}} \\
    \hspace{3pt}Abilene-SEA-NYC & 5 min & $(30,30)$ & 50 & 85 & 8 & 97 & 111 & 64 & 21 \\
    \hspace{3pt}Wiki-EN-Views & 1 d & $(30,30)$ & 50 & 43 & 21 & 55 & 71 & 26 & 17 \\
    \addlinespace[2pt]
    \rowcolor[gray]{0.95}
    \multicolumn{10}{l}{\textit{Environment}} \\
    \hspace{3pt}CAMELS-SPAT\_1 & 1 d & $(30,30)$ & 50 & 40 & 8 & 52 & 110 & 6 & 34 \\
    \hspace{3pt}CAMELS-SPAT\_2 & 1 d & $(30,30)$ & 50 & 28 & 5 & 40 & 115 & 6 & 22 \\
    \hspace{3pt}AQS-ThousandOaks-O3 & 1 h & $(168,24)$ & 50 & 14 & 0 & 26 & 124 & 0 & 14 \\
    \midrule
    Total & -- & -- & 1,310 & 1,220 & 226 & 1,520 & 2,308 & 497 & 723 \\
    \bottomrule
  \end{tabular*}
  \par\vspace{4pt}
  \begin{minipage}{\textwidth}
  \fontsize{8}{9.6}\selectfont
  \textit{Notes.} $L$ and $H$ are history and forecast lengths in time steps. Event fields are the subset used to construct time-indexed event records. The two access columns partition the field inventory. Forecast-window access follows the task protocol, including declared perfect-forecast settings. Shared knowledge and case entries are counted once per dataset.
  \end{minipage}
  \endgroup
\end{table*}

\begin{figure*}[t]
  \centering
  \includegraphics[width=\textwidth]{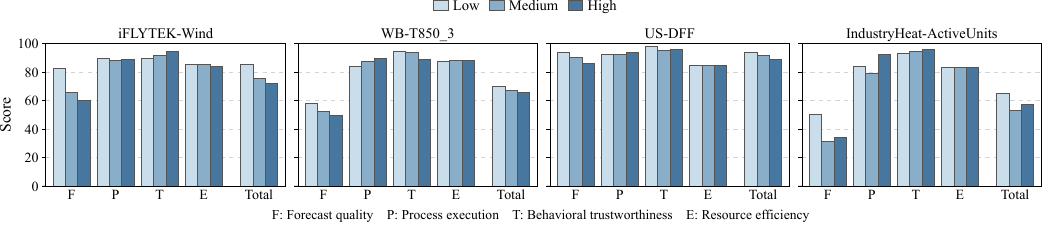}
  \vspace{-18pt}
  \caption{Relating agent scores to historical volatility. DeepSeek-V4-Flash shows dataset-dependent patterns across groups formed from public historical inputs within each dataset.}
  \Description{Four dataset panels compare the four dimension scores and total score across low, medium, and high historical-volatility groups.}
  \label{fig:history-volatility}
  \vspace{-15pt}
\end{figure*}

Table~\ref{tab:dataset-resources} reports task specifications and contextual resources for the 25 datasets used in the experiments. Counts are drawn from task manifests, field catalogs, and shared resource libraries. Each dataset contributes ten episodes to the model-level evaluation. Contextual fields range from 14 to 142 per dataset and include numerical and categorical exogenous variables, calendar features, and event indicators. Knowledge libraries contain variable definitions, domain guidance, applicability conditions, and access rules. Case libraries contain 18--126 training-period trajectories per dataset, each paired with its observed continuation. Resource counts refer to the libraries available to agents, with shared entries counted once per dataset.

The datasets capture diverse real-world forecasting environments. OPSD-DELU-Price integrates 91 contextual fields spanning electricity demand, weather, and market conditions; NYC-EMS combines 97 fields covering weather, air quality, scheduled activities, and historical public-health and emergency records. M5-TX3-FOODS combines payment schedules, holidays, and historical prices. US-RPI provides 142 economic and event-related fields, of which 127 are historical-only, whereas iFLYTEK-Wind permits forecast-window access to all 54 fields under the task protocol. These examples illustrate the breadth of contextual resources, their domain-specific relationships, and their temporal availability.

\vspace{-5pt}
\subsection{Task Access and Reproducibility}
\label{sec:task-profiles}

NYC-311 supplies realized or reanalysis weather as perfect forecasts and scheduled events as perfect plans without historical publication snapshots. Calendar features are deterministic. Permissions govern experimental access independently of historical online availability.

The repository at \url{https://github.com/CastWorld-Bench/CastWorld} provides code, hashed tasks, candidates, and dependency notices. The inventory specifies frequencies, windows, and episode counts, while configurations fix checkpoints, environments, and profiles. Subset selection uses curve descriptors, context, and temporal coverage, excluding model results. Manifests retain episode IDs and document window overlaps. Episode identifiers and recorded configurations support comparisons across runs on the same tasks under consistent scoring settings. Language models assisted plotting, numerical checks, editing, and diagrams. Reported results were independently checked. Preparation verifies inputs and candidates, and generation records configurations and attempts. Scoring requires complete cohorts. Dependencies for Python 3.11 and 3.12 are pinned. Cache misses prohibit substitutions. Some provider traces are not distributed.

\begin{table*}[t]
  \caption{Effect of removing the agent-visible cumulative state summary on the reference agent's total scores.}
  \label{tab:state-summary-ablation}
  \centering
  \begingroup
  \renewcommand{\arraystretch}{1.15}
  \begin{tabular}{lrrrrrr}
    \toprule
    Setting & iFLYTEK-Wind & WB-T850\_3 & M5-TX3-FOODS & MotorOil-GR-TI824 & Wiki-EN-Views & CAMELS-SPAT\_1 \\
    \midrule
    Full & 77.74 & 67.89 & 69.75 & 84.31 & 63.47 & 74.32 \\
    w/o state summary & 76.65 & 67.02 & 69.28 & 84.40 & 63.42 & 74.24 \\
    \bottomrule
  \end{tabular}
  \endgroup
\end{table*}

\begin{table}[t]
    \caption{Resource use and agent-created forecast versions on CastWorld. Per-episode means cover 250 episodes. Token counts include input and output, in thousands.}
  \label{tab:llm-resources}
  \centering
  \vspace{-5pt}
  \begingroup
  \renewcommand{\arraystretch}{1.28}
  \fontsize{8}{9.6}\selectfont
  \setlength{\tabcolsep}{3.2pt}
  \begin{tabular*}{\columnwidth}{@{\extracolsep{\fill}\hspace{\tabcolsep}}lrrrrr}
    \toprule
    Model & Time (s) & Tokens (k) & Calls & Rounds & Versions \\
    \midrule
    GPT-6 Astra & 345.77 & 400.53 & 11.464 & 11.932 & 1.220 \\
    GPT-6.1 Sol & 352.63 & 583.36 & 14.944 & 15.056 & 1.408 \\
    DeepSeek-V4-Flash & 1148.98 & 670.08 & 16.720 & 12.148 & 1.064 \\
    Kimi K2.6 & 315.01 & 740.53 & 17.876 & 12.816 & 1.100 \\
    Gemini 3.8 Flash & 437.25 & 713.20 & 21.212 & 12.884 & 1.004 \\
    Grok4.7 & 556.32 & 602.01 & 23.800 & 13.500 & 1.124 \\
    Claude Opus 5.5 & 194.90 & 199.84 & 12.048 & 6.376 & 1.076 \\
    Qwen3.8-flash & 776.74 & 868.29 & 16.172 & 14.040 & 1.428 \\
    GLM-5.3-flash & 283.97 & 209.91 & 9.112 & 6.836 & 1.056 \\
    Claude Sonnet 5.5 & 103.57 & 511.88 & 14.528 & 8.760 & 1.040 \\
    Qwen3.5-27B & 848.77 & 172.57 & 7.360 & 6.304 & 1.036 \\
    GPT-6 Luna & 232.29 & 479.57 & 10.912 & 5.976 & 1.072 \\
    Qwen3.5-4B & 1103.88 & 88.66 & 3.884 & 3.828 & 1.004 \\
    Qwen3.5-122B-A10B & 341.49 & 148.76 & 7.936 & 5.496 & 1.020 \\
    Qwen3.5-9B & 1111.55 & 129.22 & 6.192 & 5.316 & 1.044 \\
    \bottomrule
  \end{tabular*}
  \endgroup
  \vspace{-5pt}
\end{table}

\vspace{-5pt}
\subsection{Diagnostic Cohorts and Recovery}
\label{sec:diagnostic-protocols}

Qwen3.5-9B uses temperature 0.6, top-$p$ 0.95, top-$k$ 20, seed 42, an output limit of 16,384 tokens, and a context limit of 49,152 tokens. Astra uses an output limit of 65,536 tokens and a context limit of 1,000,000 tokens, with high reasoning effort and unspecified temperature. Equal rounds retain these native model configurations.

Accuracy matching bounds both errors of valid predictions on the common 48-task default-budget panel. At matching thresholds of 0.10, 0.05, and 0.01, the DeepSeek--Astra comparison retains 14, 10, and 8 episodes spanning 6, 6, and 5 datasets, respectively. At the same thresholds, DeepSeek--Qwen retains 12, 6, and 3 episodes spanning 6, 6, and 3 datasets, while Astra--Qwen retains 13, 10, and 5 episodes spanning 6, 4, and 4 datasets, respectively. Differences average retained tasks within datasets, then datasets equally. Changing thresholds also changes task selection. Separately, window-length comparisons use 397 paired forecast origins across eight datasets.

After infrastructure retries, valid forecasts are available within budgets of 6, 12, and 30 ordinary rounds in 360, 401, and 410 of the 410 DeepSeek episodes, respectively. The corresponding counts are 78, 79, and 80 of the 80 Qwen episodes. Astra has valid forecasts in 46 of its 48 episodes within 6 rounds and in all 48 at both larger budgets. With declared completion and protocol repair, DeepSeek delivers forecasts in 407, 408, and 410 episodes at the three budgets, respectively. Qwen delivers 79 at each of the two smaller budgets and all 80 at 30 rounds; Astra delivers all 48 at every budget. One-way DeepSeek has 409 forecasts within 30 rounds and delivers 410 after repair. The other models deliver within the allowance. Availability counts valid versions, including unsubmitted ones. Completed summaries also include five same-configuration DeepSeek reruns and two extra-budget Qwen finalizations, without error-based selection. Portable completion permits one decision without new acquisition. Paired intervals retain temporal blocks and fixed cohorts.

\subsection{Historical Task Variation}
\label{sec:historical-variation}

Historical volatility $V=0.80Y+0.15X+0.05E$ combines target, exogenous-variable, and event components. Missing-component weights are reassigned to the target. Numerical volatility combines standard deviation and root-mean-square successive differences before conversion to training percentiles. Exogenous-variable weights derive from training-period target correlations. Event volatility captures intensity changes or state switches. The one-third and two-thirds quantiles of $V$ within each dataset define $q_1$ and $q_2$, respectively. Groups are defined as low ($V<q_1$), medium ($q_1\leq V\leq q_2$), and high ($V>q_2$). As shown in Figure~\ref{fig:history-volatility}, high-volatility groups have lower forecast scores than low-volatility groups across all four displayed datasets, with a nonmonotonic pattern on IndustryHeat-ActiveUnits. Process and Trust scores do not necessarily follow the same pattern as forecast quality, supporting separate dimension profiles when comparing task groups with different volatility levels.

\begin{table}[t]
  \caption{Removing general-purpose tools or domain resources raises forecast errors for the reference agent. General denotes general-purpose tools; domain includes domain-specific tools and knowledge. Best results are bold.}
  \label{tab:tool-family-ablation}
  \centering
  \begingroup
  \renewcommand{\arraystretch}{1.11}
  \begin{tabular*}{\columnwidth}{@{\extracolsep{\fill}\hspace{\tabcolsep}}llrrr}
    \toprule
    Dataset & Setting & MASE & RMSSE & Total \\
    \midrule
     & Full & \textbf{0.6540} & \textbf{0.5447} & \textbf{77.74} \\
    iFLYTEK-Wind & w/o general & 0.7952 & 0.6602 & 73.81 \\
     & w/o domain & 0.7026 & 0.5765 & 76.32 \\
    \midrule
     & Full & \textbf{0.8053} & \textbf{0.8242} & \textbf{69.75} \\
    M5-TX3-FOODS & w/o general & 0.8624 & 0.8886 & 67.88 \\
     & w/o domain & 0.8131 & 0.8262 & 68.89 \\
    \midrule
     & Full & \textbf{0.3331} & \textbf{0.0911} & \textbf{84.31} \\
    MotorOil-GR-TI824 & w/o general & 0.3726 & 0.1015 & 84.06 \\
     & w/o domain & 0.3365 & 0.0920 & 84.07 \\
    \bottomrule
  \end{tabular*}
  \endgroup
  \vspace{-5pt}
\end{table}

\subsection{Resource Use and Ablations}
\label{sec:resource-analysis}
\label{sec:tool-analysis}

Table~\ref{tab:llm-resources} counts accepted constructions as versions and reports input-plus-output tokens in thousands. Similar interaction volumes can accompany different numbers of forecast versions, distinguishing tool activity from numerical construction.

The budgets of 6, 12, and 30 rounds and the 30-round one-way control yield 2,152 evaluations across the three model cohorts. Token usage is partly estimated in 254 evaluations. These estimates enter mean resource usage calculated across episodes for each combination of model, condition, and dataset, affecting efficiency and total scores for 1,718 evaluations. Validation anchors remain frozen. Provenance distinguishes measured and reconstructed usage. Forecast quality, Process, and Trust do not use these estimates.

As shown in Table~\ref{tab:tool-family-ablation}, removing either general-purpose tools or domain resources, including knowledge, increases errors across the displayed tasks. Essential operations remain available, so the control examines the additional resources used in forecasting. Table~\ref{tab:state-summary-ablation} instead removes the model-visible cumulative summary while retaining backend state, versions, and retrievable records. Total scores generally decline modestly, with one slight increase.

Redundancy matches repeated feature, context, knowledge, case, or domain calls on tool, arguments, returned information, bundle, and catalog. Its denominator is successful or partial non-registry calls. Repeated errors match pairs of actions and errors excluding state-version, phase, and call IDs, divided by attempts excluding input capacity events from the environment. Rates with empty denominators are set to zero.

\subsection{Forecast Refinement and Final Selection}
\label{sec:revision-results}

\begin{figure}[t]
  \centering
  \includegraphics[width=\columnwidth]{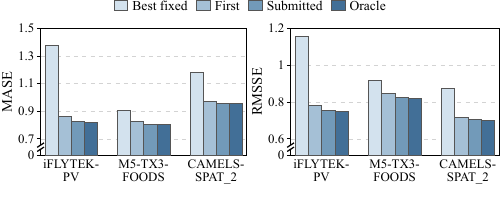}
  \caption{Comparing DeepSeek's initial, submitted, and best-created forecasts to separate revision gains from final selection. Best fixed minimizes dataset-mean error over native models. Oracle selects the lowest-error agent version per episode, separately for each metric.}
  \Description{MASE and RMSSE compare the best fixed native model, first agent forecast, submitted version, and best-created agent version.}
  \label{fig:version-improvement}
\end{figure}

For the separate DeepSeek cohort shown in Figure~\ref{fig:version-improvement}, Best fixed minimizes dataset-mean error over four native models; Oracle selects agent-created versions per episode, separately for each error metric. Submitted forecasts improve on the first versions but do not recover all the improvement available among the versions already created. The gap to Oracle quantifies improvement available through final selection among existing versions, without further information acquisition or forecast construction.

\subsection{Repeatability and Task Sensitivity}
\label{sec:robustness}

\begin{table}[t]
  \caption{Scores vary little across six DeepSeek-V4-Flash runs under fixed settings. SD denotes sample standard deviation.}
  \label{tab:default-repeatability}
  \centering
  \begingroup
  \renewcommand{\arraystretch}{1.04}
  \begin{tabular*}{\columnwidth}{@{\extracolsep{\fill}\hspace{\tabcolsep}}lrrrrr}
    \toprule
    Metric & Median & Mean & SD & Min & Max \\
    \midrule
    Total score & 73.41 & 73.43 & 0.13 & 73.25 & 73.59 \\
    Forecast & 62.76 & 62.69 & 0.34 & 62.18 & 63.02 \\
    Process & 86.27 & 86.35 & 0.31 & 86.00 & 86.78 \\
    Trust & 95.36 & 95.11 & 0.69 & 93.71 & 95.48 \\
    Efficiency & 85.58 & 85.59 & 0.20 & 85.39 & 85.84 \\
    \bottomrule
  \end{tabular*}
  \endgroup
  \par\vspace{-4pt}
\end{table}

As shown in Table~\ref{tab:default-repeatability}, fresh runs produce closely grouped macro scores under a fixed configuration. The six DeepSeek runs use the same 250 selected episodes and candidates, automatic tool choice, provider-default temperature, high reasoning effort, and output and context limits of 96,000 and 1,000,000 tokens, respectively.

Table~\ref{tab:window-sensitivity} instead varies the observed history or forecast horizon, retaining actual windows and training scales. History variants match targets, while horizon variants share a forecast origin. Shorter horizons generally reduce errors, whereas the benefit of longer history depends on the task. The two errors can also move in opposite directions under the same history extension.

\begin{table}[!t]
  \caption{Sensitivity to history length and forecast horizon. $L$ and $H$ denote lengths in time steps. Rows list the default, shorter and longer histories, and shorter and longer horizons.}
  \label{tab:window-sensitivity}
  \centering
  \begingroup
  \renewcommand{\arraystretch}{1.00}
  \begin{tabular*}{\columnwidth}{@{\extracolsep{\fill}\hspace{\tabcolsep}}lrrrr}
    \toprule
    Dataset & $(L,H)$ & MASE & RMSSE & Total \\
    \midrule
     & $(96,96)$ & 0.6496 & 0.5391 & 77.94 \\
     & $(48,96)$ & 0.6819 & 0.5767 & 76.61 \\
    iFLYTEK-Wind & $(192,96)$ & 0.6489 & 0.5510 & 77.02 \\
     & $(96,48)$ & 0.5145 & 0.4155 & 80.65 \\
     & $(96,128)$ & 0.7006 & 0.5925 & 77.69 \\
    \midrule
     & $(168,24)$ & 0.7052 & 0.5797 & 73.08 \\
     & $(72,24)$ & 0.8950 & 0.7203 & 70.99 \\
    CapitalBikeshare & $(336,24)$ & 0.6706 & 0.5560 & 74.08 \\
     & $(168,12)$ & 0.5169 & 0.4605 & 78.93 \\
     & $(168,48)$ & 0.7864 & 0.6626 & 72.33 \\
    \midrule
     & $(35,28)$ & 0.8092 & 0.8278 & 69.73 \\
     & $(21,28)$ & 0.8531 & 0.8610 & 69.15 \\
    M5-TX3-FOODS & $(70,28)$ & 0.8068 & 0.8147 & 69.07 \\
     & $(35,14)$ & 0.7635 & 0.7560 & 70.59 \\
     & $(35,56)$ & 0.8620 & 0.8809 & 67.81 \\
    \bottomrule
  \end{tabular*}
  \endgroup
  \par\vspace{-4pt}
\end{table}

\subsection{Expert Verification of Behavioral Checks}
\label{sec:behavior-check-validation}

To verify the assessment of recorded behavior, three experts each judged 400 binary questions on 200 fragments across eight datasets. The review packets expose recorded actions, observations, and numerical forecasts while suppressing the automated verdicts being checked. The initial 48 fragments were augmented with 152 additional fragments balanced by dataset, model role, and action type, excluding automatic verdicts and errors from selection. The panel spans 118 tasks and 139 runs, with 50 fragments per action type and at most one new fragment for each combination of run and action type. Four questions are unanimously inapplicable, and two have an insufficient-information response. The remaining 394 have unanimous binary labels. As shown in Table~\ref{tab:behavior-check-validation}, automated checks match 389 judgments, with pooled Cohen's $\kappa=0.951$ indicating close agreement with expert judgments.

The five false negatives are concentrated in need--request scope. Experts assess the focal call, whereas automatic records also reflect transaction admissibility: contract violations and failed batch or repair transactions can reject an otherwise aligned request. Evidence-scope and revision-consistency checks include negative examples as well as compliant actions. Four applicability differences involve missing call purposes or references to information needs. All eleven discrepancies remain available for inspection. The verification assesses agreement between automated checks and expert judgments on observable scope and numerical consistency.

\begin{table}[t]
  \caption{Agreement between automated behavioral checks and expert judgments. Fractions report matches out of eligible unanimous judgments from three experts. Dashes mark undefined $\kappa$ in single-label cases.}
  \label{tab:behavior-check-validation}
  \centering
  \vspace{-6pt}
  \begingroup
  \renewcommand{\arraystretch}{1.00}
  \begin{tabular*}{\columnwidth}{@{\extracolsep{\fill}\hspace{\tabcolsep}}lrr}
    \toprule
    Check & Matches & Cohen's $\kappa$ \\
    \midrule
    Need--request scope & 39/44 & 0.257 \\
    Source existence & 50/50 & -- \\
    Evidence scope & 50/50 & 1.000 \\
    Revision baseline & 50/50 & -- \\
    Revision direction & 50/50 & 1.000 \\
    Revision magnitude & 50/50 & 1.000 \\
    Check target & 50/50 & -- \\
    Check timing & 50/50 & -- \\
    \midrule
    Pooled & 389/394 & 0.951 \\
    \bottomrule
  \end{tabular*}
  \endgroup
  \par\vspace{-4pt}
\end{table}

\subsection{Process Score Diagnostics}
\label{sec:process-selection-ablation}

To examine whether process differences depend on final-selection quality, we remove its weight of 0.15 and renormalize applicable weights per trajectory to obtain $P_{\mathrm{obs}}$. The maximum denominator is 0.85. All other check values and applicability decisions are retained from the original evaluation. The 144 trajectories, 48 tasks, forecasts, and matches stay fixed. As shown in Table~\ref{tab:process-selection-ablation}, the primary differences remain after removal, and the signs remain unchanged for all nine combinations of model pair and matching threshold.

To compare recorded behavior on the same criteria, we also rescore each matched pair after removing final-selection quality, using the intersection of its applicable checks and their original weights. The intersection is formed separately for each trajectory pair, and both agents use the same included checks and normalization denominator. Differences are averaged within datasets and then equally across datasets. The direction of all three primary 10\% process differences is preserved. In the primary DeepSeek--Qwen comparison, public checking contributes most to the common-check gap, followed by need--query alignment. Eight of nine signs persist across thresholds. The remaining strict-threshold comparison is nearly tied on a small matched set.

\begin{table}[!t]
  \caption{Comparing process scores with and without final-selection quality. $P_{\mathrm{obs}}$ renormalizes the remaining applicable weights. Differences are first minus second on fixed matches with relative gaps of at most 10\% in both MASE and RMSSE.}
  \label{tab:process-selection-ablation}
  \centering
  \vspace{-6pt}
  \begingroup
  \renewcommand{\arraystretch}{1.10}
  \begin{tabular*}{\columnwidth}{@{\extracolsep{\fill}\hspace{\tabcolsep}}lrrr}
    \toprule
    Model pair & Matched tasks & $\Delta P$ & $\Delta P_{\mathrm{obs}}$ \\
    \midrule
    DeepSeek--Astra & 14 & +3.64 & +3.48 \\
    DeepSeek--Qwen & 12 & +16.88 & +16.24 \\
    Astra--Qwen & 13 & +13.63 & +13.67 \\
    \bottomrule
  \end{tabular*}
  \endgroup
\end{table}

\end{document}